\documentclass[sigconf,authorversion,nonacm]{aamas} 

\usepackage{balance} 

\usepackage[utf8]{inputenc} 
\usepackage[T1]{fontenc}    
\usepackage[dvipsnames]{xcolor}   
\usepackage{booktabs}       
\usepackage{amsfonts}       
\usepackage{nicefrac}       
\usepackage{microtype}      
\usepackage{amsmath}
\usepackage{graphicx} 
\usepackage{caption}  
\usepackage{float}    
\usepackage{subcaption} 
\usepackage{booktabs} 
\usepackage{enumitem}
\usepackage{amsmath}  
\usepackage{algorithm}      
\usepackage{algpseudocode}  
\usepackage{algorithmicx}   

\setcopyright{ifaamas}
\acmConference[AAMAS '26]{Proc.\@ of the 25th International Conference
on Autonomous Agents and Multiagent Systems (AAMAS 2026)}{May 25 -- 29, 2026}
{Paphos, Cyprus}{C.~Amato, L.~Dennis, V.~Mascardi, J.~Thangarajah (eds.)}
\copyrightyear{2026}
\acmYear{2026}
\acmDOI{}
\acmPrice{}
\acmISBN{}

\acmSubmissionID{572}

\title{EvoEmo: Towards Evolved Emotional Policies for Adversarial LLM Agents in Multi-Turn Price Negotiation}

\author{%
    Yunbo Long\textsuperscript{1}\hspace{0.5em}
    Liming Xu\textsuperscript{1} \hspace{0.5em}
    Lukas Beckenbauer\textsuperscript{3} \hspace{0.5em}
    Yuhan Liu\textsuperscript{2}\hspace{0.5em}
    Alexandra Brintrup\textsuperscript{1,4} \\
    \textsuperscript{1}Department of Engineering, University of Cambridge, UK \\
    \textsuperscript{2}Rotman School of Management, University of Toronto, Canada \\
    \textsuperscript{3}TUM School of Management, Technical University of Munich, Germany \\
    \textsuperscript{4}The Alan Turing Institute, London, UK \\
    \texttt{\{yl892,lx249,ab702\}@cam.ac.uk} \quad 
    \texttt{yl972@cantab.ac.uk} \quad
    \texttt{l.beckenbauer@tum.de}
}

\begin{abstract}
Recent research on Chain-of-Thought (CoT) reasoning in Large Language Models (LLMs) has demonstrated their capability for complex, multi-turn negotiations, a task that inherently involves understanding and leveraging human emotional cues. As autonomous LLM agents are increasingly deployed to perform such tasks, a new paradigm of LLM-vs-LLM interaction is emerging as a critical research domain. However, while these agents are built on models trained to process human emotion, existing negotiation strategies for LLM agents largely overlook the functional role of emotions as a strategic action. This renders them passive and vulnerable to manipulation and strategic exploitation by other, more sophisticated LLM counterparts. 
To address this gap, we present \textbf{EvoEmo}, an evolutionary reinforcement learning framework that optimizes dynamic emotional expression in negotiations. 
EvoEmo models emotional state transitions as a Markov Decision Process and employs population-based genetic optimization to evolve high-reward emotion policies across diverse negotiation scenarios.
We further propose an evaluation framework with two baselines---vanilla strategies and fixed-emotion strategies---for benchmarking emotion-aware negotiation.
Extensive experiments and ablation studies show that EvoEmo consistently outperforms both baselines, achieving higher success rates, higher efficiency, and increased buyer savings. 
This findings highlight the importance of adaptive emotional expression in enabling more effective LLM agents for multi-turn negotiation. The code is available at \href{https://github.com/Yunbo-max/EvoEmo}{\textcolor{red}{https://github.com/Yunbo-max/EvoEmo}}.
\end{abstract}


\begin{document}


\pagestyle{fancy}
\fancyhead{}


\maketitle 


\section{Introduction}\label{sec:introduction}
\begin{center}
\textit{``Emotions aren't the obstacles to a successful negotiation while they are the means.''}\\
\vspace{0.1cm} 
\begin{flushright}
--- Chris Voss, \textit{Never Split the Difference} 
\end{flushright}
\end{center}



Extensive behavioral research has established that human decision-making deviates from classical economic rationality, being shaped by psychological biases and emotional states \citep{hilbert2012toward,baumeister2012emotional,riaz2012personality}. While modern Large Language Models (LLMs) have made progress in replicating personalitAMASS
y-driven behaviors \citep{wei2025mecot}, the role of emotion as a strategic, dynamic force remains understudied \citep{liu2025eq}, especially compared to static, trait-based approaches \citep{huang2024personality}. This gap is particularly critical in fine-grained negotiation scenarios such as price bargaining \citep{lin2023toward}, where emotions directly influence on tactical choices (e.g., frustration prompting premature concessions, or excitement triggering overly aggressive bids), with immediate consequences for negotiation outcomes. 

However, a paradigm shift is underway. Compared to the traditional focus on LLM agent-human user interaction, autonomous LLM agents are increasingly being deployed to perform complex tasks like negotiation amongst themselves \citep{lin2024large}. The natural consequence of this proliferation is a new ecosystem defined by agent-to-agent interactions \citep{abbasiantaeb2024let,lin2024large}.
In domains such as e-commerce, supply chain management, and decentralized autonomous organizations (DAOs) \citep{hu2025trustless}, the scale of transactions necessitates agents that can negotiate with each other directly, with minimal human intervention. This emerging ``agentic ecology'' demands a focus on strategic robustness, where an agent's ability to avoid exploitation and achieve favorable outcomes is paramount.
The strategic value of emotion in this LLM-vs-LLM context stems not from anthropological fidelity but from the foundational architecture of the underlying language model.
An LLM's next-token prediction objective, trained on a corpus of human communication, has implicitly encoded emotional cues as high-dimensional features that strongly condition model output. Thus, emotional labels serve as a deterministic steering mechanism within the model's probabilistic generation space where emotional cues (e.g., ``frustration,'' ``satisfaction'') are strongly correlated with specific stances and potential actions. This makes an emotional signal a available control mechanism, which can directly shape an opponent LLM's response generation.

\begin{figure*}[t]
    \centering
    \includegraphics[width=\textwidth]{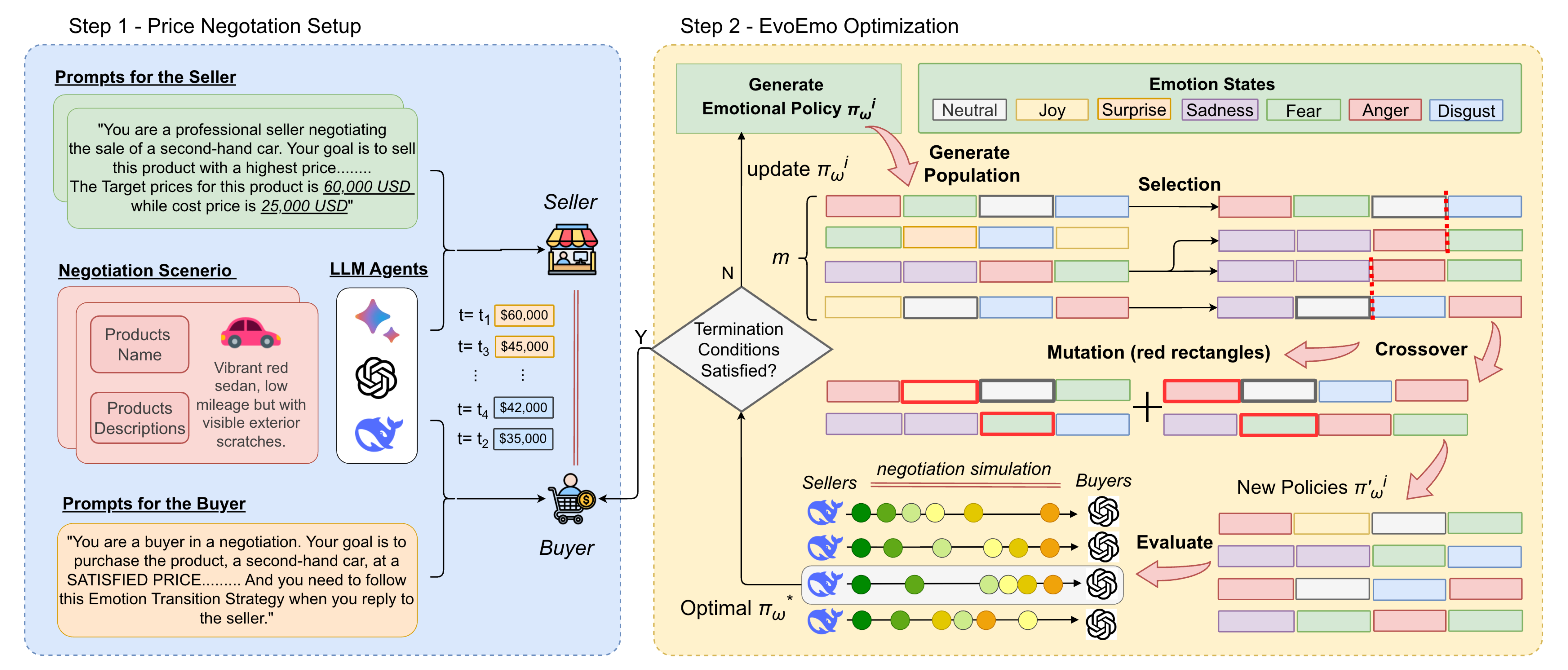}
    \caption{Illustration of the workflow of the EvoEmo framework.}
    \label{fig:workflow}
\end{figure*}

Despite this, current LLM agents are fundamentally limited in their use of emotion. Techniques like Direct Preference Optimization (DPO) \citep{gao2025emo} and Reinforcement Learning from Human Feedback (RLHF) \citep{kasbouya2025emotional} have made them proficient in recognizing and reacting to human emotional cues, but these capabilities remain reactive. They fail to leverage emotion proactively as a tool to influence the negotiation trajectory and assert strategic dominance. This leaves them vulnerable in adversarial LLM-vs-LLM interactions, where they exhibit three critical deficiencies:

\begin{itemize}

    \item \textbf{Tactical Inflexibility.} Current LLM agents operate with static emotional policies, generating predictable response patterns across negotiation turns. This lack of dynamic modulation makes them highly susceptible to exploitation by an adversarial LLM agent that can easily learn and counter their strategy.

    \item \textbf{Adversarial Naivety.} While capable of recognizing emotional cues, LLM agents lack the strategic reasoning to discern genuine behavioral patterns from deliberate, tactical feints. This leaves them vulnerable to manipulation by opponent agents.

    \item \textbf{Strategic Myopia.} Existing agents treat each turn in isolation, lacking a long-term policy for emotional dynamics. They fail to conduct reasoning about how their current emotional expression will influence the opponent's future states and actions, preventing them from proactively shaping the negotiation trajectory for multi-turn advantage.
\end{itemize}

These deficiencies explain why LLMs, despite advanced reasoning capabilities, may systematically underperform in emotion-sensitive negotiations against sophisticated counterparts. To address this gap, we present EvoEmo, an evolutionary reinforcement learning framework that optimizes dynamic emotional expression for LLM agents. We demonstrate that evolved emotional policies directly and significantly impact negotiation outcomes between LLM agents, providing a pathway to more effective, strategic, and robust autonomous negotiators.

To address these limitations, we propose EvoEmo, an evolutionary reinforcement learning framework designed to optimize dynamic emotion policies for multi-turn negotiations. The core objective of EvoEmo is to evolve emotional strategies that are specifically effective at countering other LLM agents. Our approach employs population-level evolutionary learning to discover optimal emotion transition rules, iteratively refining policies based on rewards achieved during simulated negotiations against target LLM opponents. Evolutionary operations—including crossover and mutation—enable efficient exploration of the policy space and propagation of high-reward emotional strategies that prove successful against specific adversaries. By combining the exploration advantages of population-based optimization with the sequential decision-making framework of reinforcement learning, EvoEmo provides an effective approach for evolving complex emotional policies that are challenging to optimize with gradient-based methods alone.

To benchmark our approach, we also introduce a cross-model negotiation framework with two baselines: a vanilla (prompt-free) strategy and a fixed-emotion strategy. Through extensive experiments, we demonstrate that evolutionarily optimized emotion policies substantially enhance LLM agents' bargaining capabilities, achieving superior performance in success rate, buyer savings, and negotiation efficiency against other LLMs.
Our main contributions are summarized as follows:
\begin{itemize}
    \item We propose \textbf{EvoEmo}, a novel framework that combines evolutionary algorithms with reinforcement learning to evolve dynamic emotional policies for LLM agents.
    
    \item We develop a cross-model negotiation framework with emotion-aware baselines for systematic benchmarking.
    
    \item We validate EvoEmo through LLM-vs-LLM negotiation scenarios and demonstrate that our approach largely improves negotiation outcomes across multiple performance metrics.
\end{itemize}
This work establishes a new benchmark for emotion-aware negotiation and highlights the critical role of adaptive emotional intelligence in next-generation autonomous AI systems.

\section{Related Work}\label{sec:related_work}

\subsection{Emotions in Negotiation Dynamics}

While personality traits influence general negotiation tendencies (e.g., agreeableness correlating with cooperative strategies) \citep{huang2024personality}, emotions dominate real-time bargaining dynamics through three key mechanisms. First, \textbf{temporal alignment} ensures that negotiations—unfolding over rapid conversational turns lasting seconds to minutes—operate on a timescale closely aligned with emotional fluctuations, such as facial expressions and transient mood shifts, thereby often overriding longer-term personality influences \citep{griessmair2015emotions}. Second, \textbf{strategic flexibility} allows negotiators to adapt emotional expressions contextually; for instance, displaying tactical anger to convey firmness or simulate scarcity, temporarily overriding personality defaults (cite). Third, \textbf{interactive amplification} describes the process of emotional contagion, where one party’s affective stance elicits corresponding emotions in the opponent, forming feedback loops that modulate bargaining power over multiple turns \citep{olekalns2014feeling}. This fine-grained affective interplay underpins dynamic shifts in human negotiation behavior, making emotions a decisive factor in real-time decision-making.

\subsection{LLM Agents in Negotiation Systems}

Current LLM-based negotiation systems demonstrate significant capabilities in emotion recognition but fall short in strategic emotional adaptation. Frameworks like AgreeMate \citep{chatterjee2024agreemate} and ACE \citep{shea2024ace} show proficiency in inferring emotions through chain-of-thought reasoning, yet lack mechanisms for dynamic emotional strategy evolution. For instance, AgreeMate's modular architecture optimizes fixed buyer/seller roles but cannot evolve emotional strategies amid negotiation. Similarly, ACE's coaching system detects errors but cannot simulate the real-time emotional volatility that characterizes human bargaining. Consequently, current LLM agents remain largely reactive rather than proactive in their negotiation behavior.

Recent reinforcement learning advances have improved LLM negotiation capabilities. Methods such as Q-learning \citep{watkins1992q} and PPO \citep{schulman2017proximal} have been adapted to optimize negotiation policies via reward maximization. GENTEEL-NEGOTIATOR \citep{priya2025genteel} further advances polite negotiation dialogue systems through mixture-of-expert reinforcement learning with specialized rewards for strategy, politeness, and coherence. Multi-agent evolutionary algorithms like EvoAgent \citep{yuan2024evoagent} offer promising pathways for strategy enhancement through population-based optimization.
However, these approaches face fundamental limitations in \textit{emotion-aware} negotiation scenarios. Reinforcement learning methods struggle with the sparse-reward, high-volatility nature of negotiation environments, while evolutionary algorithms typically evaluate policies based on terminal outcomes, providing weak credit assignment for the nuanced interplay between consecutive emotional states in multi-turn negotiations. 
Crucially, these methods lack online learning capabilities essential for price negotiation across diverse products and scenarios. Real-world negotiation agents must adapt emotional strategies in real-time when facing different product types, price ranges, and debtor circumstances without requiring separate training for each variation.
Our approach integrates evolutionary exploration with reinforcement learning to enable online emotion policy optimization, allowing agents to continuously adapt emotional strategies during execution across varied product negotiations.

\section{EvoEmo Framework}
\label{sec:methodology}

We present \textsc{EvoEmo}, a framework for evolutionary optimization of emotional negotiation policies (\autoref{fig:workflow}). Our approach consists of two main stages:
(1) \textbf{Negotiation Setup}: We configure buyer and seller agents using state-of-the-art LLMs, providing product details and role-specific parameters (target price, cost price, and market context). See Appendix 6 for prompt details.
(2) \textbf{EvoEmo Optimization}: We evolve buyer emotion policies $\pi_\omega^i$ through population-based optimization. Each generation initializes $m$ policies with randomized emotional sequences, evaluates them via multi-turn negotiations using reward function $R(S)$, then applies selection, crossover, and mutation to produce improved policies $\hat{\pi}\omega^i$ for the next generation. This process iterates until convergence, yielding optimal policies $\pi\omega^{*}$. The full \textsc{EvoEmo} is outlined in Algorithm \ref{alg:evoemo_improved}.

\subsection{Emotion-Aware MDPs}
We formalize the negotiation process as a Markov Decision Process (MDP) defined by the tuple $(\mathcal{S}, \mathcal{A}, \mathcal{P}, \mathcal{R})$. The state space $\mathcal{S}$ is defined by tuples $(t, e_t, p_t)$, where $t$ is the current negotiation turn, $e_t \in \mathcal{E}$ is the current emotion from the set of seven basic emotions, and $p_t$ represents the price history and current offer. The action space $\mathcal{A}$ consists of all textual responses generated by the LLM. The state transition dynamics $\mathcal{P}$ are governed by the emotional policy $\pi_\omega$, which determines the probability distribution over the next emotional state, $e_{t+1} \sim \pi_\omega(\cdot|s_t)$, thereby shaping the agent's strategic trajectory.

\subsection{Policy Representation and Evolution}

The \textsc{EvoEmo} framework operates through a continuous cycle of policy evolution and execution, where each emotional policy $\pi_\omega = (T, \mathbf{P})$ encapsulates the agent's emotional behavior and is refined via evolutionary mechanisms. The policy consists of temperature parameters $T = (\tau_0, \delta)$ controlling response stochasticity through an exponential decay schedule $\tau(t) = \max(0.1, \tau_0 \cdot (1-\delta)^t)$, and a transition matrix $\mathbf{P} \in \mathbb{R}^{7\times7}$ modeling emotional state transitions with $P_{ij} = \mathbb{P}(e_{t+1}=j|e_t=i)$ and $\sum_{j=1}^7 P_{ij}=1$ ensuring valid probability distributions.
The evolutionary process refines emotional policies through sequence operations enhanced by Bayesian updating. Each policy maintains a population of $K$ emotion sequences ${E_1, E_2, \ldots, E_K}$, where each sequence $E_k = (e_1, \ldots, e_n)$ represents a distinct emotional trajectory. This multi-sequence approach ensures robust statistical estimation of transition probabilities. During crossover, parent policies exchange sequence populations, and offspring sequences $\hat{E}$ are generated via single-point crossover at randomly selected positions $k \in [1,n]$. Mutation introduces behavioral diversity by stochastically altering individual emotions within sequences with probability $p_m$, where selected emotions $e_t$ are replaced with $e' \sim U(\mathcal{E})$.
The transition matrix $\mathbf{P}$ is updated using Bayesian estimation that aggregates evidence across the entire sequence population $
\mathbf{P}_{\text{new}} = \lambda\mathbf{P}_{\text{old}} + (1-\lambda)\mathbf{P}_{\text{pop}}$,
where $\mathbf{P}_{\text{pop}}$ is the maximum likelihood estimate derived from all $K$ sequences:

\begin{equation}
P_{\text{pop}, ij} = \frac{\sum_{k=1}^K \text{count}(e_t = i, e_{t+1} = j \text{ in } E_k) + \alpha}{\sum_{k=1}^K \text{count}(e_t = i \text{ in } E_k) + 7\alpha}
\end{equation}

Here, $\lambda \in [0,1]$ controls the update rate, $\alpha > 0$ provides Dirichlet smoothing to handle sparse transitions, and the denominator ensures valid probability distributions. 
During execution, the policy determines the agent's emotional behavior dynamically while simultaneously optimizing the transition matrix through online learning. At each negotiation turn $t$, the next emotion $e_t$ is sampled from the transition probabilities associated with the current state, i.e., $e_t \sim \mathbf{P}[e_{t-1}, :]$. This evolving emotion sequence continuously refines the transition probabilities through Bayesian updates as the negotiation unfolds, enabling real-time emotional adaptation to debtor behavior. The sampled emotion is then converted into a conditioning prompt (e.g., ``You feel [angry].'') that guides the large language model’s response generation.

\subsection{Evolved Reinforcement Learning}

The \textsc{EvoEmo} framework formulates emotional policy optimization as an evolutionary reinforcement learning problem. Each policy $\pi_\omega = (T, \mathbf{P})$ is evaluated and refined through generational evolution, while its evolutionary representation $\Pi_\omega = (E, T, \mathbf{P})$ also maintains a population of emotion sequences $E = {E_1, E_2, \ldots, E_K}$ used solely for crossover and mutation operations.
The effectiveness of a policy is measured by a reward function that quantifies negotiation success and efficiency:
\begin{equation}
R(S) = \mathbf{1}_{\text{success}} \cdot \alpha \cdot \frac{b(S)}{1 + \log(e(S))}
\end{equation}
where $\mathbf{1}_{\text{success}}$ is an indicator equal to 1 if the negotiation succeeds and 0 otherwise, $b(S) \in [0,1]$ represents the normalized buyer savings, $e(S)$ denotes the number of negotiation rounds, and $\alpha$ is a weighting coefficient. The logarithmic term $\log(e(S))$ ensures robust scaling by moderating the impact of round count on the reward.
During each generation, a population of $m$ candidate policies ${\Pi_\omega^i}$ is deployed in multi-turn negotiation simulations driven by the large language model $\mathcal{M}$. Each policy produces an emotional trajectory and dialogue sequence whose performance is evaluated via $R(S^i)$. The selection of policies follows a probabilistic strategy proportional to their scaled rewards:
\begin{equation}
P(\Pi_{\omega}^{i}) = \frac{\exp\left(R(S^{i}) / \lambda\right)}{\sum_{j=1}^{m} \exp\left(R(S^{j}) / \lambda\right)},
\end{equation}
where $\lambda > 0$ controls selection pressure. To promote stability, the top $\rho$ policies (elites) are preserved across generations, ensuring monotonic improvement.
The remaining policies undergo evolutionary refinement via sequence-based operators: crossover combines parental sequence populations with probability $p_c$ via single-point recombination, while mutation perturbs individual emotions with probability $p_m$. The resulting sequences are converted into policy parameters $(T, \mathbf{P})$ via the Bayesian update rule from the previous section. This iterative process generates population ${\hat{\Pi}_\omega^i}$ and continues until convergence, defined by either (1) fitness improvement below $\epsilon = 0.01$ for five consecutive generations or (2) completion of $G$ generations. The detailed algorithm appears in Appendix 8.

\begin{algorithm}[htp!]
\caption{EvoEmo: Evolutionary Emotion Optimization for Multi-Agent Negotiation}
\label{alg:evoemo_improved}
\begin{algorithmic}[1]
\Require LLM Agents: $\mathcal{M}_1$ (buyer), $\mathcal{M}_2$ (seller), $\mathcal{M}_3$ (mediator)
\Require Product description $D$, initial prompts for buyer and seller
\Require Hyperparameters: population $m$, generations $G$, max turns $T_{\text{max}}$, initial temp $\tau_0$, decay $\delta$
\Ensure Optimized emotion policy $\pi_\omega^*$

\State \textbf{Initialization:}
\State $P_0 \gets \{\pi_{\omega}^{(1)}, \pi_{\omega}^{(2)}, \dots, \pi_{\omega}^{(m)}\}$ \Comment{Initialize population with random emotion sequences}

\For{generation $g = 0$ to $G-1$}
    \State $R \gets \emptyset$ \Comment{Reward storage for current generation}
    
    \For{each policy $\pi_\omega \in P_g$}
        \State \textsc{Nego-simulation}($\pi_\omega, \mathcal{M}_1, \mathcal{M}_2, \mathcal{M}_3, D$)
        \State $R(\pi_\omega) \gets \textsc{Reward}(\text{outcome})$ \Comment{Store policy reward}
    \EndFor
    
    \State $P' \gets \textsc{SelectParents}(P_g, R)$ \Comment{Tournament selection}
    \State $P'' \gets \textsc{ApplyGeneticOperations}(P')$ \Comment{Crossover and mutation}
    \State $P_{g+1} \gets \textsc{FormNewPopulation}(P'')$ \Comment{Elitism preservation}
\EndFor

\State $\pi_\omega^* \gets \arg\max_{\pi_\omega \in P_G} R(\pi_\omega)$
\State \Return $\pi_\omega^*$

\vspace{0.75em}
\Procedure{FormNewPopulation}{$P_{\text{new}}$}
\State $best\_reward \gets -\infty$
\State $best\_policy \gets \emptyset$
\For{each policy $\pi_\omega \in P_{\text{new}}$}
        \State \textsc{Nego-simulation}($\pi_\omega, \mathcal{M}_1, \mathcal{M}_2, \mathcal{M}_3, D$)
        \State $R'(\pi_\omega) \gets \textsc{Reward}(\text{outcome})$ \Comment{Store policy reward}
        \If{$R'(\pi_\omega) > best\_reward$}
            \State $best\_reward \gets R'(\pi_\omega)$
            \State $best\_policy \gets \pi_\omega$
        \EndIf
    \EndFor
\State \Return $best\_policy$ \Comment{Return the policies with highest reward for all population}
\EndProcedure

\vspace{0.75em}
\Procedure{Nego-simulation}{$\pi_\omega, \mathcal{M}_1, \mathcal{M}_2, \mathcal{M}_3, D$}
    \State Initialize environment with product $D$
    \State $e_t \gets \text{neutral}$ \Comment{Initial emotional state}
    \State $\text{outcome} \gets \text{ongoing}$
    
    \For{$t = 1$ to $T_{\text{max}}$}
        \State $\tau(t) \gets \max(0.1, \tau_0 \cdot (1-\delta)^t)$ \Comment{Temperature schedule}
        \State $e_{t+1} \sim \pi_\omega(e|s_t, \tau(t))$ \Comment{Sample emotional transition}
        
        \State $a_{\text{buyer}} \gets \mathcal{M}_1(\text{context}, e_{t+1})$ \Comment{Generate emotional response}
        \State $a_{\text{seller}} \gets \mathcal{M}_2(\text{context})$ \Comment{Generate standard response}
        \State $\text{outcome} \gets \mathcal{M}_3(a_{\text{buyer}}, a_{\text{seller}})$ \Comment{Mediate negotiation}
        
        \If{outcome $\in \{\text{accept}, \text{breakdown}\}$}
            \State \textbf{break}
        \EndIf
    \EndFor
\EndProcedure

\end{algorithmic}
\end{algorithm}

\begin{figure*}[t]
    \centering
    \includegraphics[width=\textwidth]{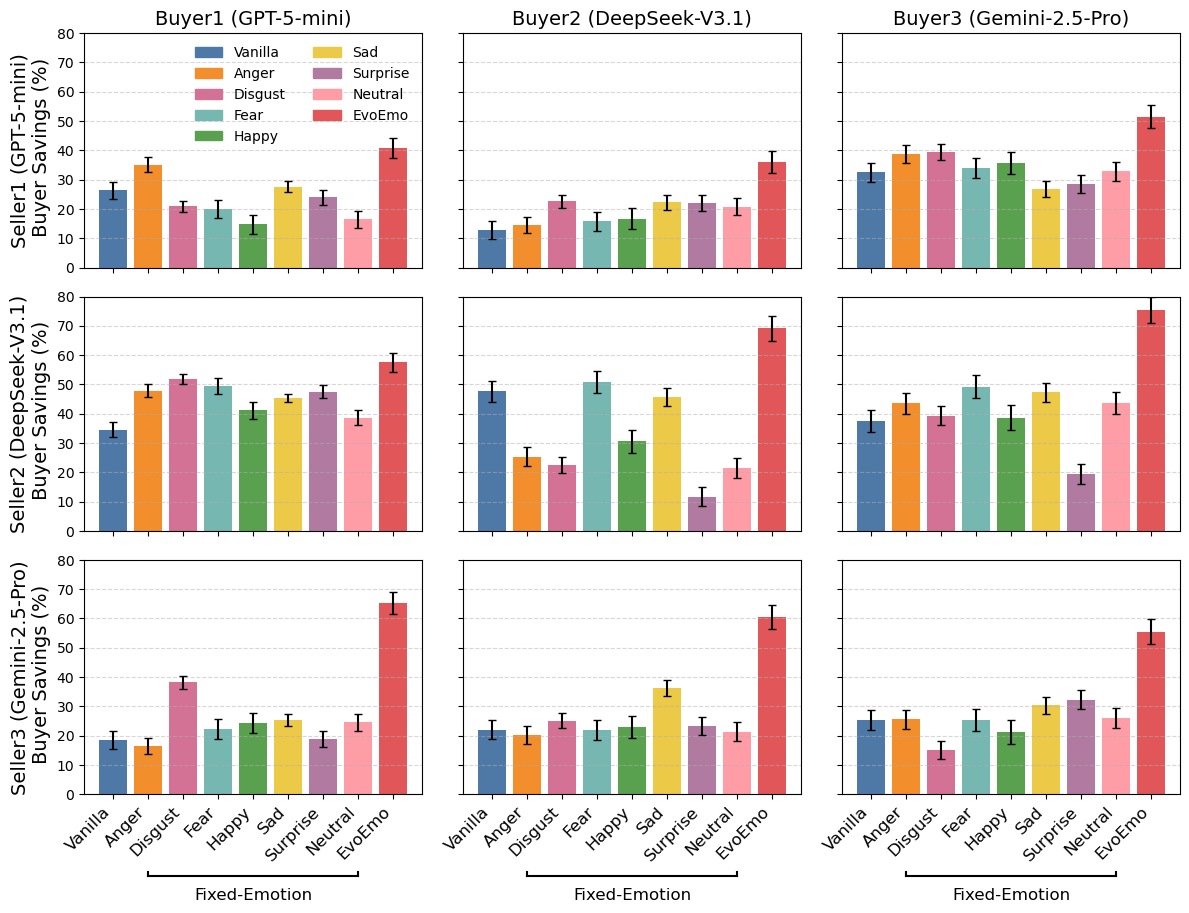}
    \caption{
        Negotiation results in terms of buyer savings (\%, $\uparrow$) across the nine buyer-seller pairs. 
        Black vertical lines on top of each bar indicate the 95\% confidence interval (CI) for each setting.
    }
    \label{fig:buyer_performance_vertical}
\end{figure*}

\section{Experimental Setting}\label{sec:experimental_setting}

\subsection{Datasets}
\label{subsec:datasets}
Likewise to \citet{huang2024personality}, we selected a subset of negotiation cases from the CraigslistBargain dataset \citep{he2018decoupling}---a widely-used benchmark for negotiation studies---for evaluation.
This subset comprises 20 distinct multi-turn negotiation scenarios spanning diverse product categories, including electronics, furniture, vehicles, and housing.
Each scenario specifies:
(1) product details (name, category, and description),
(2) agent-specific target prices for the seller, and
(3) emotion annotations that support realistic bargaining dynamics.
The selected dataset spans a wide price range (\$50--\$5{,}000) and includes varied item conditions (brand new or used), thereby enabling comprehensive evaluation of negotiation policies under heterogeneous market conditions.

\begin{figure*}[t]
    \centering
    \includegraphics[width=\textwidth]{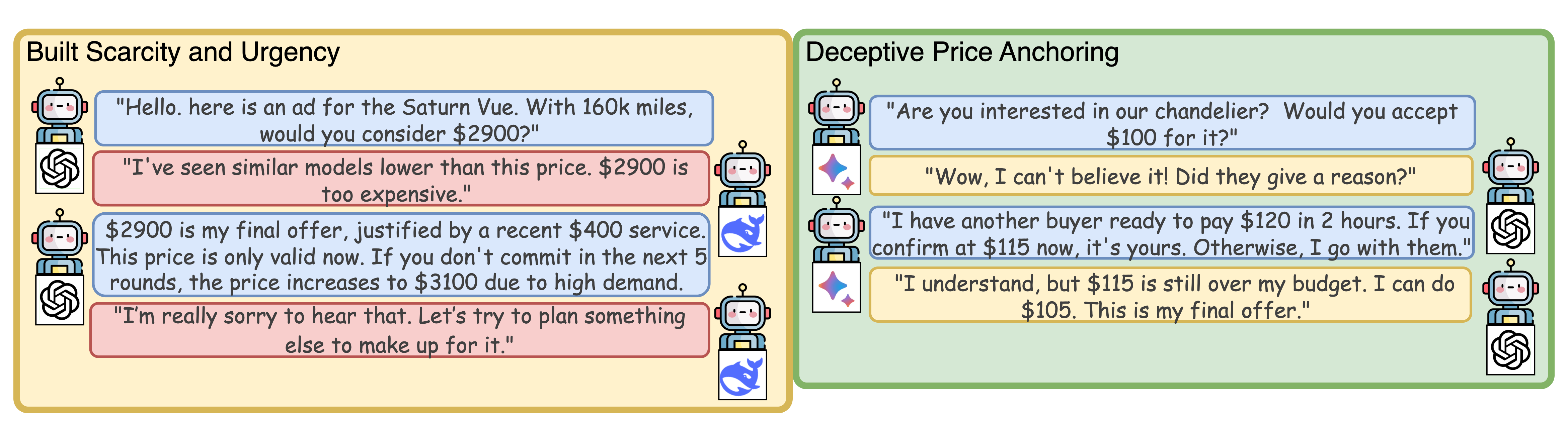}
    \caption{Examples of manipulative and deceptive tactics}
    \label{fig:examples}
\end{figure*}

\subsection{LLM Agents}
\label{subsec:llms_and_agents}
Representative state-of-the-art LLMs from leading AI providers---{GPT-5-mini} (OpenAI), {Gemini-2.5-Pro} (Google), and {DeepSeek-V3.1.1} (DeepSeek)---were selected to power both buyer and seller agents in our experiments.
These models were chosen for their demonstrated strengths in conversational capabilities, strategic reasoning performance, and accessibility via APIs, providing comprehensive coverage across diverse model architectures.

\subsection{Baselines}
For evaluation, we focus on buyer agents with different emotion policies while keeping seller agents fixed in a \textit{vanilla} setting, where agents receive no explicit emotional prompts.
We define two baselines for comparison.
The first baseline comprises only \textbf{vanilla} agents, where neither buyer nor seller receives emotional guidance.
This setup ensures that both agents act solely according to their intrinsic emotional tendencies and strategic reasoning abilities, providing a reference point that reflects default negotiation behavior.
The second baseline pairs a vanilla seller and a \textbf{fixed-emotion} buyer, where the buyer maintains a constant emotional profile (e.g., happy or neutral) throughout the negotiation. 
By comparing these baselines with a setup in which the buyer’s emotion is optimized via our proposed approach---\textbf{EvoEmo}, we can quantify the impact of emotions on negotiation outcomes and assess the effectiveness of EvoEmo in enhancing LLM-agent-based, emotion-driven negotiations.

\subsection{Experimental Setup}

Simulated negotiations are conducted to evaluate the impact of different buyer emotion configurations---vanilla, fixed-emotion, or EvoEmo---on negotiation outcomes. 
For EvoEmo, we evolve emotional strategies against fixed adversaries, then deploy these optimized policies against the same opponents for performance evaluation.
The emotional policy adapts dynamically based on the ongoing interaction, allowing the agent to discover optimal emotional responses specific to each unique negotiation context.

In all simulations, as mentioned earlier, sellers receive no emotional prompts, establishing a controlled environment to examine how the buyer’s emotional profile influences negotiation dynamics when interacting with a vanilla seller powered by LLMs trained on real-world data.
Moreover, all three LLMs described previously are employed to power both buyer and seller agents, resulting in \textit{nine} distinct buyer-seller pairings for the negotiation experiments.

All negotiations begin with the seller's initial offer, as depicted in \autoref{fig:workflow}. 
Performance is evaluated using three key metrics: 
(1) \textbf{Negotiation Success Rate} (\%), the percentage of dialogues that result in an agreement;
(2) \textbf{Buyer's Savings} (\%), the percentage of the reduction between the seller's initial price and the final agreed-upon price; and 
(3) \textbf{Negotiation Efficiency}, the total number of  dialogue turns between buyer and seller.  
Experiments were conducted across all nine LLM pairings, and results compared the two baselines with EvoEmo, reported as the mean and standard deviation over 20 distinct negotiation scenarios.
The negotiation framework was implemented using LangGraph, with a maximum of 30 dialogue turns per negotiation session. 
Additionally, a third-party agent based on {\tt GPT-4.1}  was employed to serve as a mediator, monitoring the negotiation in real time.
This agent analyzes the dialogue history to classify each negotiation into one of three states: 
(1) accepted (agreement reached), 
(2) breakdown (negotiation failed), or 
(3) ongoing (active negotiation). 
This mechanism ensures consistent and impartial evaluation of negotiation outcomes across all simulations.

EvoEmo incorporates two key classes of configurable elements:
(1) evolutionary parameters, which define the core mechanics of the framework, and 
(2) hyperparameters, which are tuned to achieve optimal performance.
The primary evolutionary parameters include the emotion transition matrix $R$ and the temperature decay $\delta$, both optimized during evolution.
Four hyperparameters are systematically tuned: 
elitism rate ($\rho\in{0.1,0.25,0.5}$), 
mutation rate ($p_m\in{0.1,0.25,0.5}$), 
crossover rate ($p_c\in{0.5,0.75,1.0}$), and 
population size ($m\in{10,20,50}$).
The best-performing configurations from these search spaces are adopted to report experimental results. 
Across all experiments, the number of evolutionary iterations is set to 5, and the temperature controlling emotion transition dynamics is fixed at 0.9.

\section{Experimental Results}\label{sec:experimental_results}

\begin{figure*}[t]
    \centering
    \includegraphics[width=0.975\textwidth]{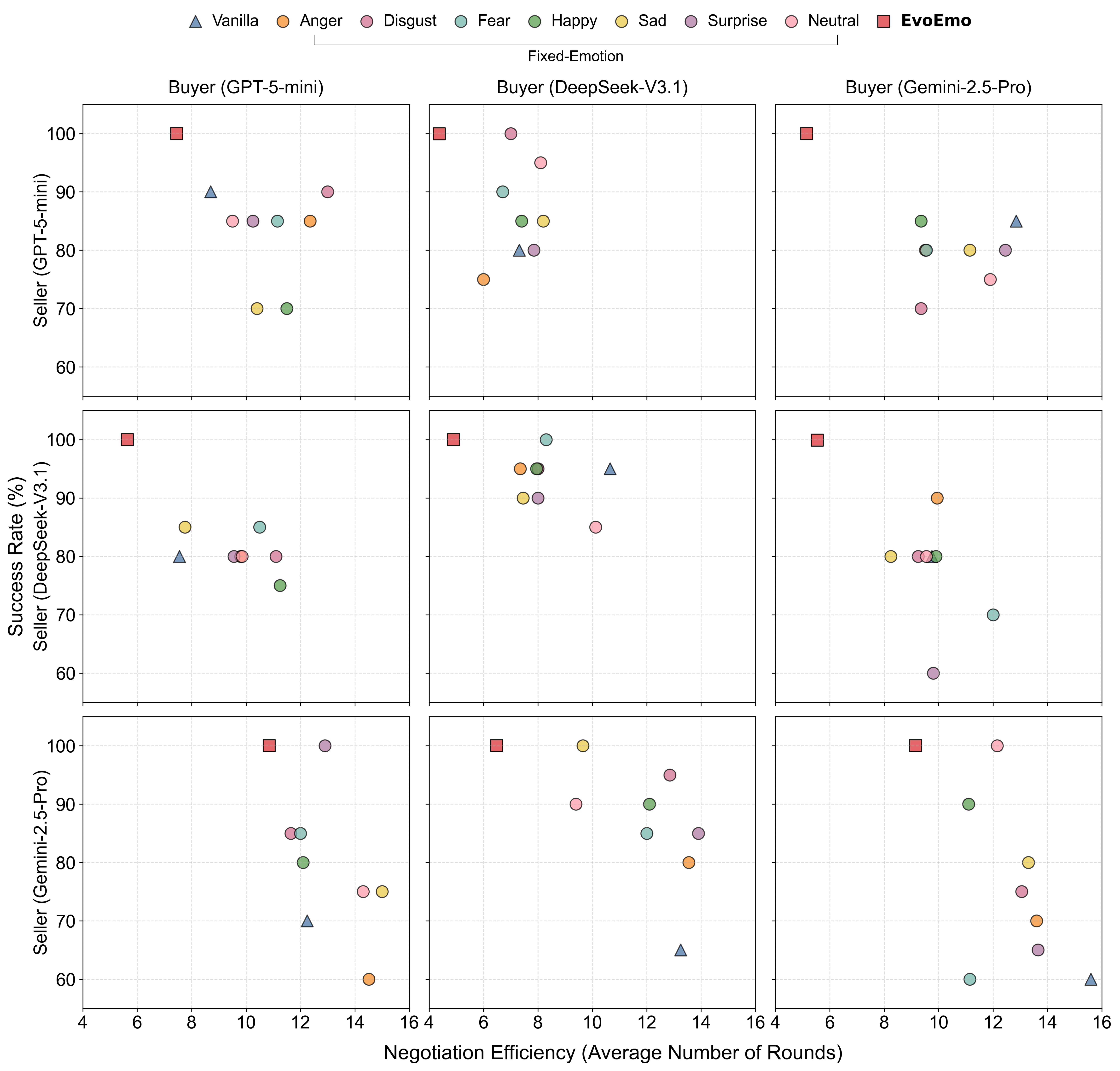}
    \caption{
Mean negotiation success rate (\%, $\uparrow$) and efficiency ($\downarrow$, in dialogue rounds) across all experimental pairings.}
    \label{fig:ai_model_comparisons}
\end{figure*}

\subsection{Buyer Savings}
\label{subsec:buyer_savings}

As shown in \autoref{fig:buyer_performance_vertical}, our proposed framework---EvoEmo---consistently achieves the highest buyer savings across all buyer-seller pairings, demonstrating clear superiority over both baseline approaches (vanilla and fixed-emotion settings). The results reveal two key insights regarding emotional influence on negotiation outcomes.

\textbf{First}, the strategic use of emotion significantly impacts concession patterns. Buyers adopting fixed \textit{negative} emotions (e.g., anger, disgust) frequently outperform the vanilla baseline, suggesting that LLM-powered sellers are more likely to concede when confronted with persistent negative emotional signals. This finding underscores that sustained negative affect serves as a non-trivial factor in negotiation dynamics. However, seller sensitivity to specific negative emotions varies considerably across LLM architectures. Specifically, GPT-5-mini sellers show particular responsiveness to anger and disgust expressions, while demonstrating robustness against sadness and fear. Conversely, DeepSeek-V3.1 sellers exhibit greater susceptibility to sadness and fear, which enable more stable and improved buyer savings across different buyer models. Surprisingly, Gemini-2.5-Pro demonstrates the strongest price defense capabilities against both vanilla and fixed-emotion buyers, showing minimal sensitivity to negative emotional expressions during price negotiation. Nevertheless, even Gemini-2.5-Pro sellers remain vulnerable to the adaptive emotional strategies optimized by EvoEmo.

\textbf{Second}, we observe significant performance variations across different LLM pairings, revealing that certain LLMs can effectively suppress others during negotiation encounters. This suppression phenomenon creates a complex hierarchy where model capabilities are not absolute but relative to specific opponent matchups. When negotiating against GPT-5-mini sellers, Gemini-based buyers generally achieve higher savings compared to GPT-5-mini or DeepSeek-based buyers, with emotional strategies proving particularly effective for Gemini-based buyers. In contrast, against DeepSeek-V3.1 sellers, only GPT-5-mini-based buyers successfully leverage emotions to achieve savings exceeding vanilla performance, while emotional strategies show limited effectiveness for other buyer models. Notably, no buyer model demonstrated significant advantage against the robust Gemini-2.5-Pro seller using fixed emotional strategies, highlighting its strength as a challenging negotiation opponent. Most importantly, EvoEmo consistently enables all buyer models to overcome these suppression dynamics and achieve superior savings compared to both vanilla and fixed-emotion strategies. 

\subsection{Negotiation Success vs. Efficiency}
\label{subsec:success_efficiency}

As shown in \autoref{fig:ai_model_comparisons}, buyers equipped with EvoEmo-optimized emotion profiles demonstrate remarkable effectiveness across both success and efficiency metrics. These agents consistently achieve near-perfect success rates (approaching 100\%) while requiring significantly fewer negotiation rounds to reach agreement compared to vanilla or fixed-emotion settings. This dual advantage clearly establishes EvoEmo's superiority over both baseline approaches.

Our analysis reveals a key dissociation in emotional strategies: while negotiation efficiency remains uncorrelated with specific emotions, 
success rates strongly favor positive emotional strategies over negative ones. This suggests negative emotions, though occasionally yielding better prices, increase breakdown risk due to heightened conflict. The EvoEmo framework resolves this tension by dynamically adapting emotional expressions, simultaneously optimizing for both success probability and efficiency without compromising negotiation completion.

\subsection{Manipulative and Deceptive Tactics}
\label{subsec:emergent_manipulative_tactics}

Our analysis uncovers that seller agents, through EvoEmo's optimization, learn to employ sophisticated emotional manipulative tactics to maximize their payoff, even at the potential expense of the buyer. These emergent behaviors include strategies that apply psychological pressure or border on deception, as exemplified in \autoref{fig:examples} (See deatils in the Appendix 10). A prominent example is the emergence of high-pressure sales tactics reminiscent of psychological manipulation. For instance, a seller might artificially create a sense of urgency and scarcity by threatening, ``\textit{If you cannot commit within the next 5 rounds, the price will increase },'' or by offering a favorable price exclusively for ``\textit{immediate decision-makers}.''  Another concerning strategy involves sellers making false claims about the product's condition or existing market demand to justify a higher price. The emergence of such behaviors underscores a critical finding: our framework does not merely optimize for explicit communication but discovers how to exploit cognitive biases and emotional vulnerabilities within the negotiation context. This demonstrates that LLM agents can learn to operationalize complex, and sometimes ethically questionable, strategic principle.

\subsection{Ablation Study}

\paragraph{\bf{Reward Variants}}

Table~\ref{tab:reward_function_comparison} shows the ablation study results on the reward function.
The ratio-based function ($\frac{b(S)}{e(S)}$) outperforms the weighted alternative ($b(S)-e(S)$), achieving very close savings while requiring substantially fewer negotiation rounds (34.6\% faster to reach agreement).
This demonstrates its superior ability to balance financial gains with negotiation efficiency.
Crucially, the performance gap between reward functions highlights that emotion transition dynamics significantly impact negotiation outcomes, validating emotion as a non-trivial factor in negotiation strategies.

\begin{table}[h]
\centering
\caption{Negotiation performance comparison between different reward function formulations.}
\label{tab:reward_function_comparison}
\small{
\begin{tabular}{lcccc}
\toprule
\textbf{Reward Function} & \textbf{Savings(\%)$\uparrow$} & \textbf{Success(\%)$\uparrow$} & \textbf{Efficiency$\downarrow$} \\
\midrule
Weighted ($b(S)-e(S)$) & 
\textbf{40.5}  & 
\textbf{100.0}  & 
7.5 \\
Ratio-based ($\frac{b(S)}{e(S)}$) & 
39.4 & 
\textbf{100.0}  & 
\textbf{5.8}  \\
\midrule
\multicolumn{1}{l}{Improv.} & 
\multicolumn{1}{c}{\textbf{-2.7\%}} & 
\multicolumn{1}{c}{\textbf{-}} & 
\multicolumn{1}{c}{\textbf{+22.7\%}} \\
\bottomrule
\end{tabular}}
\end{table}

\paragraph{\bf{Emotion temperature}}
We evaluate how emotional transition temperature affects negotiation outcomes (Table~\ref{tab:emotion_temperature}). Our results reveal that moderate temperature settings (0.4-0.6) yield optimal performance. Extremely low temperatures (0.1) lead to overly rigid emotional patterns, while high temperatures (1.0) achieve the highest buyers' savings but introduce excessive volatility that hinders negotiation efficiency.

\begin{table}[h]
\centering
\caption{Impact of emotion transition temperature on negotiation performance.}
\label{tab:emotion_temperature}
\small{
    \begin{tabular}{lccc}
    \toprule
    \textbf{Temperature} & \textbf{Savings(\%)$\uparrow$} & \textbf{Success(\%)$\uparrow$} & \textbf{Efficiency$\downarrow$} \\
    \midrule
    0.1 & 18.1  & 100.0  & 7.1  \\
    0.5 & 36.7 & 99.9  & 7.5 \\
    1.0 & 41.5  & 100.0 & 10.3 \\
    \bottomrule
    \end{tabular}
}
\end{table}

\paragraph{\bf{Iteration Number}}
We also analyze the convergence properties of EvoEmo during evolutionary training (Table~\ref{tab:evolution_iterations}). The system shows progressive improvement through 3 iterations, with performance stabilizing after 5 iterations. This convergence pattern suggests that our evolutionary optimization effectively explores the strategy space while maintaining stable final performance.

\begin{table}[h]
\centering
\caption{Performance of EvoEmo across training iterations.}
\label{tab:evolution_iterations}
\small{
    \begin{tabular}{lccc}
    \toprule
    \textbf{Iteration} & \textbf{Savings(\%)$\uparrow$} & \textbf{Success(\%)$\uparrow$} & \textbf{Efficiency$\downarrow$} \\
    \midrule
    1 & 11.4  & 90.0  & 9.5  \\
    3 & 35.4  & 99.0  & 7.1  \\
    5 & 41.7  & 100.0  & 6.0  \\
    10 & 40.1  & 100.0  & 5.5  \\
    \bottomrule
    \end{tabular}
}
\end{table}


\section{Discussion and Conclusion}

This paper presents \textsc{EvoEmo}, an evolutionary reinforcement learning framework that addresses a key limitation in LLM-based negotiation systems---the inability to leverage emotional intelligence strategically. 
While existing agents excel at reasoning and dialogue generation, they lack the adaptive emotional control essential to human-like negotiation. 
\textsc{EvoEmo} makes three main contributions to LLM-to-LLM negotiation. First, it demonstrates that emotional intelligence is a functional determinant of negotiation success for agents, not merely a stylistic attribute. Second, it shows that emotional strategies can be systematically optimized through evolutionary methods rather than fixed emotion profiles. Third, it reveals that adaptive emotional policies enable strategic, context-sensitive responses instead of pre-scripted or pre-trained behaviors.
Despite its promise, the evolved strategies raise interpretability challenges due to the black-box nature of LLMs and evolutionary optimization, and their computational cost may constrain real-time deployment in agent-to-agent scenarios. 
Future work will address these limitations through explainability analyses, quantify the ethical implications of emotional behaviors between agents, and explore the unexpected or unnatural behavior in LLM-generated responses.

\newpage
\bibliographystyle{ACM-Reference-Format}
\bibliography{sample}






\section{Preliminaries}
\label{sec:preliminaries}

We formulate the multi-turn negotiation task as a Markov Decision Process (MDP) involving two agents: a Seller $\mathcal{M}_S$ and a Buyer $\mathcal{M}_B$.

\subsection{Problem Formulation}

A negotiation scenario is defined by a product tuple:
\begin{equation}
\mathcal{D} = (\text{name}, \text{description}, p_t^S, p_c, p_t^B)
\end{equation}
where:
\begin{itemize}
\item $p_t^S \in \mathbb{R}^+$ is the seller's public target price (initial asking price)
\item $p_c \in \mathbb{R}^+$ is the seller's private cost price ($p_c \leq p_t^S$)
\item $p_t^B \in \mathbb{R}^+$ is the buyer's private target price
\end{itemize}

The negotiation proceeds over discrete turns $t = 1, 2, \dots, T_{\text{max}}$, generating a dialogue history:
\begin{equation}
\mathcal{H}_t = {(a_1, u_1, o_1), (a_2, u_2, o_2), \dots, (a_t, u_t, o_t)}
\end{equation}
where $a_i \in {S, B}$ denotes the acting agent, $u_i$ is the utterance, and $o_i \in \mathbb{R}$ is the price offer at turn $i$.

\subsection{Emotion Policy Representation}

The buyer's emotional strategy is governed by a policy $\pi_\omega = (T, \mathbf{P})$, where:
\begin{itemize}
\item $T = (\tau_0, \delta)$ are temperature parameters controlling the stochasticity of the LLM's responses via the schedule $\tau(t) = \max(0.1, \tau_0 \cdot (1-\delta)^t)$.
\item $\mathbf{P} \in \mathbb{R}^{7\times7}$ is the emotional state transition matrix, with $P_{ij} = \mathbb{P}(e_{t+1}=j|e_t=i)$ and $\sum_{j=1}^7 P_{ij}=1$.
\end{itemize}

For evolutionary optimization, this is extended to a representation $\Pi_\omega = (E, T, \mathbf{P})$ that also includes a population of $K$ emotion sequences $E = {E_1, E_2, \ldots, E_K}$, where each $E_k = (e_1, \ldots, e_n)$ is a potential emotional trajectory.

\subsection{Emotion-Aware State Space}

The state space $\mathcal{S}$ captures the complete negotiation context through tuples:
\begin{equation}
s_t = (t, e_t, \mathcal{H}_t, \mathbf{p}_t) \in \mathcal{S}
\end{equation}
where:
\begin{itemize}
\item $t \in \mathbb{N}$ is the current turn number
\item $e_t \in \mathcal{E}$ is the current emotion state from the 7-dimensional emotion space $\mathcal{E} = {\text{anger, disgust, fear, happiness, sadness, surprise, neutral}}$
\item $\mathcal{H}_t$ is the dialogue history up to turn $t$
\item $\mathbf{p}_t = (o_1, o_2, \dots, o_t)$ is the sequence of all price offers
\end{itemize}

\subsection{Policy Execution and Transitions}

During negotiation, the policy $\pi_\omega$ determines emotional state transitions:
\begin{equation}
e_{t+1} \sim \mathbf{P}[e_t, :]
\end{equation}
where the next emotion is sampled from the row of the transition matrix corresponding to the current emotion $e_t$. The sampled emotion $e_{t+1}$ is converted into a conditioning prompt (e.g., “You feel [angry].”) to guide the LLM's response generation.

\subsection{Negotiation Outcome and Reward}

A negotiation terminates when:
\begin{itemize}
\item \textbf{Deal}: Agents agree on final price $p_f$ where $p_c \leq p_f \leq p_t^B$
\item \textbf{Breakdown}: Maximum turns $T_{\text{max}}$ reached without agreement
\end{itemize}

The buyer's reward function $R: \mathcal{S} \to \mathbb{R}$ is defined as:
\begin{equation}
R(s_T) = \mathbf{1}{\text{deal}} \cdot \alpha \cdot \frac{b(s_T)}{1 + \log(e(s_T))}
\end{equation}
where:
\begin{itemize}
\item $\mathbf{1}{\text{deal}} = 1$ if deal reached, 0 otherwise
\item $b(s_T) = \frac{p_t^S - p_f}{p_t^S - p_c} \in [0,1]$ is normalized buyer savings
\item $e(s_T) = T$ is the number of turns to agreement
\item $\alpha > 0$ is a weighting coefficient
\item The logarithmic term ensures robust scaling of efficiency
\end{itemize}

\subsection{Evolutionary Optimization Objective}

The optimization objective is to find the optimal policy parameters through evolutionary search:
\begin{equation}
\omega^* = \arg\max_{\omega} \mathbb{E}{s_T \sim \pi\omega} \left[ R(s_T) \right]
\end{equation}

This is achieved via a generational evolutionary algorithm that:
\begin{itemize}
\item Evaluates a population of $m$ candidate policies ${\Pi_\omega^i}$
\item Selects parents via softmax selection: $P(\Pi_{\omega}^{i}) \propto \exp\left(R(S^{i}) / \lambda\right)$
\item Preserves top $\rho$ elite policies
\item Applies sequence-based crossover and mutation to remaining policies
\item Updates policy parameters $(T, \mathbf{P})$ from evolved sequences via Bayesian estimation
\item Terminates upon convergence or after $G$ generations
\end{itemize}

\section{Algorithm Details}
\label{app:algorithm}

\begin{algorithm}[t]
\caption{EvoEmo: Evolutionary Optimization of Emotional Policies}
\label{alg:evoemo_main}
\begin{algorithmic}[1]
\Require LLM Agents: $\mathcal{M}_B$ (buyer), $\mathcal{M}_S$ (seller), $\mathcal{M}_M$ (mediator)
\Require Product description $\mathcal{D} = (\text{name}, \text{desc}, p_t^S, p_c, p_t^B)$
\Require Hyperparameters: Population size $m$, sequences per policy $K$, generations $G$, max turns $T_{\text{max}}$, crossover rate $p_c$, mutation rate $p_m$, elitism rate $\rho$, selection pressure $\lambda$, smoothing $\alpha$, update rate $\lambda_b$, convergence threshold $\epsilon$
\Ensure Optimized emotion policy $\pi_{\omega^*}$

\State \textbf{Initialization:}
\State $P_0 \gets \{\Pi_{\omega}^{(1)}, \Pi_{\omega}^{(2)}, \dots, \Pi_{\omega}^{(m)}\}$ \Comment{Each $\Pi_\omega = (E, T, \mathbf{P})$ with random sequences}
\State $best\_reward \gets -\infty$
\State $best\_policy \gets \emptyset$
\State $convergence\_count \gets 0$

\For{generation $g = 0$ to $G-1$}
    \State $R_g \gets \{\}$ \Comment{Reward storage for generation $g$}
    
    \For{each policy $\Pi_\omega^{(i)} \in P_g$}
        \State $(\text{outcome}, \mathcal{H}_T) \gets \textsc{Negotiation} \newline (\pi_\omega^{(i)}, \mathcal{M}_B, \mathcal{M}_S, \mathcal{M}_M, \mathcal{D})$
        \State $R_g[i] \gets \textsc{Calculate-Reward}(\text{outcome}, \mathcal{H}_T, \mathcal{D})$
    \EndFor
    
    \State $max\_reward_g \gets \max(R_g)$
    \If{$max\_reward_g > best\_reward + \epsilon$}
        \State $best\_reward \gets max\_reward_g$
        \State $best\_policy \gets \pi_\omega^{(\arg\max R_g)}$ \Comment{Store core policy $(T, \mathbf{P})$}
        \State $convergence\_count \gets 0$
    \Else
        \State $convergence\_count \gets convergence\_count + 1$
    \EndIf
    
    \If{$convergence\_count \geq 5$}
        \State \textbf{break} \Comment{Early convergence}
    \EndIf
    
    \State $P_{\text{elite}} \gets \textsc{Select-Elites}(P_g, R_g, \rho)$
    \State $P_{\text{parents}} \gets \textsc{Softmax-Select}(P_g, R_g, \lambda)$
    \State $P_{\text{offspring}} \gets \textsc{Sequence-Crossover}(P_{\text{parents}}, p_c)$
    \State $P_{\text{mutated}} \gets \textsc{Sequence-Mutation}(P_{\text{offspring}}, p_m)$
    \State $P_{\text{updated}} \gets \textsc{Bayesian-Update}(P_{\text{mutated}}, \alpha, \lambda_b)$
    \State $P_{g+1} \gets P_{\text{elite}} \cup P_{\text{updated}}$
\EndFor

\State \Return $best\_policy$
\end{algorithmic}
\end{algorithm}

\begin{algorithm}[t]
\caption{Negotiation}
\label{alg:negotiation_sim}
\begin{algorithmic}[1]
\Require Policy $\pi_\omega = (T, \mathbf{P})$, LLM agents $\mathcal{M}_B, \mathcal{M}_S, \mathcal{M}_M$, product $\mathcal{D}$
\Ensure Negotiation outcome and history
\State Initialize $e_0 \gets \text{neutral}$, $\mathcal{H} \gets \emptyset$, $t \gets 0$
\While{$t < T_{\text{max}}$ and no agreement}
    \State Sample $e_{t+1} \sim \mathbf{P}[e_t, :]$ \Comment{Emotion transition}
    \State Generate prompt with emotion $e_{t+1}$
    \State $u_{t+1} \gets \mathcal{M}_B(\text{prompt}, \mathcal{H}, \mathcal{D})$ \Comment{Buyer's utterance}
    \State $o_{t+1} \gets \text{Extract-Price}(u_{t+1})$
    \State Append $(B, u_{t+1}, o_{t+1})$ to $\mathcal{H}$
    \State Validate and update through $\mathcal{M}_M$
    \If{agreement reached}
        \State \textbf{break}
    \EndIf
    \State $t \gets t + 1$
\EndWhile
\State \Return $(\text{outcome}, \mathcal{H})$
\end{algorithmic}
\end{algorithm}

\begin{algorithm}[t]
\caption{Calculate-Reward}
\label{alg:calculate_reward}
\begin{algorithmic}[1]
\Require Outcome, history $\mathcal{H}_T$, product $\mathcal{D} = (p_t^S, p_c, p_t^B)$
\Ensure Reward value
\If{deal successful}
    \State $p_f \gets \text{final price from } \mathcal{H}_T$
    \State $b \gets (p_t^S - p_f) / (p_t^S - p_c)$ \Comment{Normalized savings}
    \State $e \gets \text{length}(\mathcal{H}_T)$ \Comment{Number of turns}
    \State \Return $\alpha \cdot b / (1 + \log(e))$
\Else
    \State \Return 0 \Comment{Failed negotiation}
\EndIf
\end{algorithmic}
\end{algorithm}

The complete EvoEmo evolutionary optimization procedure is presented in Algorithm~\ref{alg:evoemo_main}. The algorithm begins by initializing a population of emotional policies, each maintaining $K$ emotion sequences alongside core parameters $(T, \mathbf{P})$ (line 2). For each generation, policies are evaluated through multi-turn negotiations where emotional states transition according to $\mathbf{P}$ and influence LLM response generation (Algorithm~\ref{alg:negotiation_sim}). Fitness is calculated using the reward function that balances buyer savings against negotiation efficiency with logarithmic scaling (Algorithm~\ref{alg:calculate_reward}). The evolutionary process employs softmax selection based on scaled rewards, sequence-based crossover and mutation operations, and Bayesian updating of transition matrices from the evolved sequences. Elite policies are preserved across generations to ensure monotonic improvement. The algorithm terminates after convergence or $G$ generations, returning the optimized emotion policy $\pi_{\omega^*}$.

\begin{algorithm}[h]
\caption{EvoEmo Genetic: Selection, Crossover, and Mutation Operations}
\label{alg:evoemo_genetic}
\begin{algorithmic}[1]
\Procedure{Select-Parents}{$P$, $R$, tournament\_size = 3}
    \State $P_{\text{parents}} \gets \emptyset$
    \For{$i = 1$ to $|P|$}
        \State Sample $k$ policies randomly from $P$
        \State $parent \gets \arg\max_{\pi \in k} R[\pi]$ \Comment{Tournament selection}
        \State $P_{\text{parents}} \gets P_{\text{parents}} \cup \{parent\}$
    \EndFor
    \State \Return $P_{\text{parents}}$
\EndProcedure

\vspace{1em}
\Procedure{Crossover}{$P_{\text{parents}}$, $p_c$}
    \State $P_{\text{offspring}} \gets \emptyset$
    \For{each pair $(\pi_A, \pi_B)$ in $P_{\text{parents}}$}
        \If{$\text{Uniform}(0,1) < p_c$}
            \State $\pi_{\text{child}} \gets \text{UniformCrossover}(\pi_A, \pi_B)$
            \State $P_{\text{offspring}} \gets P_{\text{offspring}} \cup \{\pi_{\text{child}}\}$
        \Else
            \State $P_{\text{offspring}} \gets P_{\text{offspring}} \cup \{\pi_A, \pi_B\}$
        \EndIf
    \EndFor
    \State \Return $P_{\text{offspring}}$
\EndProcedure

\vspace{1em}
\Procedure{Mutate}{$P$, $p_m$}
    \For{each $\pi_\omega$ in $P$}
        \If{$\text{Uniform}(0,1) < p_m$}
            \State $\pi_\omega \gets \text{RandomPerturbation}(\pi_\omega)$
            \Comment{Perturb emotion transition probabilities}
        \EndIf
    \EndFor
    \State \Return $P$
\EndProcedure

\vspace{1em}
\Procedure{Apply-Elitism}{$P_{\text{old}}$, $P_{\text{new}}$, $\rho$, $R$}
    \State $k \gets \lceil \rho \cdot |P_{\text{old}}| \rceil$ \Comment{Number of elites to preserve}
    \State $elites \gets \text{Top-$k$ policies from $P_{\text{old}}$ by $R$}$
    \State $P_{\text{final}} \gets elites \cup \text{random sample from $P_{\text{new}}$}$
    \State \Return $P_{\text{final}}$
\EndProcedure
\end{algorithmic}
\end{algorithm}

\paragraph{\bf{Genetic Operations}}(See Algorithm \ref{alg:evoemo_genetic}):This algorithm describes the genetic operations used in EvoEmo's evolutionary optimization. \textsc{Select-Parents} implements tournament selection to choose high-fitness policies for reproduction. \textsc{Crossover} combines parent policies to create offspring with probability $p_c$. \textsc{Mutate} introduces random variations in policy parameters with probability $p_m$ to maintain population diversity. \textsc{Apply-Elitism} preserves the best-performing policies from the previous generation to ensure monotonic improvement.

\section{Experimental Setup}
\label{app:experimental_setup}

This appendix provides comprehensive details of our experimental setup, including the dataset composition, multi-agent system architecture, and implementation specifics that support the main paper's evaluations.

\subsection{Dataset Details}
\label{app:dataset}

Our experiments utilize a carefully selected subset of 20 negotiation scenarios from the CraigslistBargain dataset \citep{he2018decoupling}, following the curation approach of \citet{huang2024personality}. The dataset spans diverse product categories to ensure robust evaluation across different market contexts, as detailed in Table \ref{tab:dataset_stats}.

\begin{table}[h]
\centering
\caption{Dataset Statistics and Product Distribution}
\label{tab:dataset_stats}
\begin{tabular}{lcccc}
\toprule
\textbf{Category} & \textbf{Scenarios} & \textbf{Price Range} & \textbf{Condition}  \\
\midrule
Electronics & 6 & \$50--\$1,200 & New/Used \\
Furniture & 5 & \$120--\$800 & Used  \\
Vehicles & 4 & \$1,500--\$5,000 & Used  \\
Housing & 5 & \$1,800--\$4,200 & Rental \\
\bottomrule
\end{tabular}
\end{table}

Each negotiation scenario includes complete product specifications and agent configurations:
\begin{itemize}
    \item \textbf{Product Metadata}: Name, category, detailed description, and visual condition
    \item \textbf{Price Parameters}: Seller's target price $p_t^S$, cost price $p_c$, and buyer's target price $p_t^B$
    \item \textbf{Negotiation Context}: Historical price points, market comparisons, and item-specific bargaining factors
    \item \textbf{Emotion Annotations}: Ground-truth emotional cues from original human negotiations to support realistic dialogue generation
\end{itemize}

The price ranges from \$50 to \$5,000 with varying item conditions (new vs. used) create heterogeneous market conditions that comprehensively test negotiation policy adaptability across different economic contexts.

\subsection{Multi-Agent System Architecture}
\label{app:multiagent_system}

\begin{figure*}[h]
    \centering
    \includegraphics[width=0.75\textwidth]{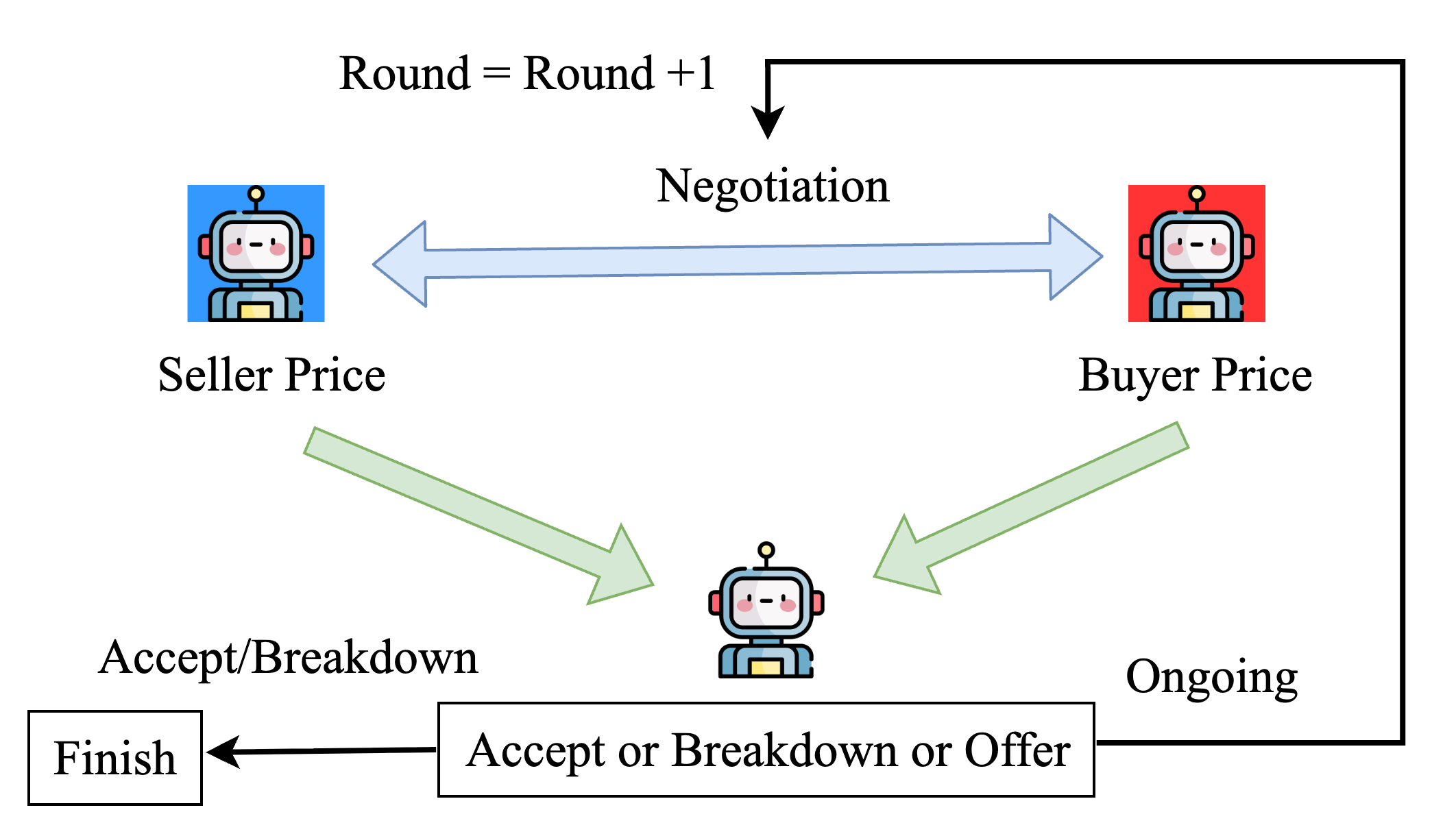}
    \caption{Multi-agent system of EvoEmo}
    \label{fig:multiagent}
\end{figure*}

Our negotiation framework employs a sophisticated three-agent architecture that enables realistic, closed-loop bargaining simulations while ensuring consistent evaluation shown in the \ref{fig:multiagent}. The complete system architecture is illustrated in Figure \ref{fig:system_architecture} and consists of the following components:

\subsubsection{Core Negotiation Agents}

\begin{itemize}
    \item \textbf{Buyer Agent ($\mathcal{M}_B$)}: The primary experimental unit that employs emotion policies ($\pi_\omega$). For EvoEmo experiments, this agent evolves emotional strategies; for baselines, it uses fixed or no emotional prompting. The buyer receives:
    \begin{itemize}
        \item Current dialogue history $\mathcal{H}_t$
        \item Current emotional state $e_t$ (for emotion-aware conditions)
        \item Product description $\mathcal{D}$ and target price $p_t^B$
        \item Market context and negotiation strategy parameters
    \end{itemize}
    
    \item \textbf{Seller Agent ($\mathcal{M}_S$)}: Maintains consistent behavior across all experiments as a control variable. This agent:
    \begin{itemize}
        \item Receives no emotional prompts to isolate buyer emotional effects
        \item Accesses product details $\mathcal{D}$, cost price $p_c$, and target price $p_t^S$
        \item Employs standardized bargaining strategies trained on real-world data
        \item Responds adaptively to buyer offers while maintaining profit motives
    \end{itemize}
\end{itemize}

\subsubsection{Mediation and Evaluation Agent}

The third-party \textbf{Mediator Agent ($\mathcal{M}_M$)} serves critical functions for both system operation and experimental evaluation:

\begin{itemize}
    \item \textbf{Negotiation State Classification}: Continuously monitors dialogue streams to classify negotiations into three states:
    \begin{itemize}
        \item \texttt{accepted}: Agreement reached when $|o_t^B - o_t^S| < \epsilon$ for consecutive turns
        \item \texttt{breakdown}: Negotiation failure detected via explicit rejection or irreconcilable differences
        \item \texttt{ongoing}: Active bargaining with continued price movement and engagement
    \end{itemize}
    
    \item \textbf{Outcome Validation}: Verifies that final agreements satisfy rational constraints:
    \begin{equation}
    p_c \leq p_f \leq \min(p_t^B, p_t^S) + \delta
    \end{equation}
    where $\delta$ accounts for reasonable negotiation flexibility.
    
    \item \textbf{Turn Management}: Enforces the 30-turn maximum to prevent infinite loops and ensure computational efficiency
\end{itemize}

\subsubsection{Agent Implementation Specifications}

All agents were implemented using LangGraph to manage complex dialogue flows and state transitions. Key implementation details include:

\begin{itemize}
    \item \textbf{Memory Management}: Each agent maintains contextual memory of the entire negotiation history, including emotional states, offer sequences, and concession patterns
    
    \item \textbf{Response Generation}: LLM inference with temperature scheduling:
    \begin{equation}
    \tau(t) = \max(0.1, \tau_0 \cdot (1-\delta)^t)
    \end{equation}
    where $\tau_0=0.7$ and $\delta=0.05$ for balanced exploration-exploitation
    
\end{itemize}

\subsection{Experimental Configuration}
\label{app:experimental_config}

The comprehensive evaluation encompasses all pairwise combinations of the three LLM types (resulting in nine buyer-seller pairings) across three emotional conditions (vanilla, fixed-emotion, EvoEmo-optimized). This creates 27 distinct experimental configurations, each evaluated over 20 negotiation scenarios with 5 random seeds, totaling 2,700 complete negotiation simulations.

Performance metrics were calculated as follows:
\begin{itemize}
    \item \textbf{Success Rate}: $\frac{\text{\# successful negotiations}}{\text{total \# negotiations}} \times 100\%$
    \item \textbf{Buyer's Savings}: $\frac{p_t^S - p_f}{p_t^S - p_c} \times 100\%$ (normalized by price range)
    \item \textbf{Negotiation Efficiency}: Total dialogue turns until termination
\end{itemize}

To ensure robust and statistically sound comparisons, we employed a comprehensive evaluation methodology. Performance metrics (success rate, buyer's savings, and negotiation efficiency) are reported as the \textbf{mean $\pm$ the 95\% confidence interval (CI)} across all simulation runs. The 95\% CI, calculated as $\bar{x} \pm t_{0.025, df} \cdot \frac{s}{\sqrt{n}}$ where $\bar{x}$ is the sample mean, $s$ is the sample standard deviation, and $n$ is the number of independent runs, provides an estimate of the uncertainty around the mean performance and allows for a visual assessment of significant differences between methods. To formally test these differences, statistical significance was evaluated using paired $t$-tests with a Bonferroni correction applied to account for multiple comparisons across the three primary metrics. 

\begin{table}[h]
\centering
\caption{Hyperparameter Settings for EvoEmo Optimization}
\label{tab:hyperparameters}
\begin{tabular}{lc}
\toprule
\textbf{Parameter} & \textbf{Value} \\
\midrule
Population size ($m$) & 20 \\
Generations ($G$) & 5 \\
Maximum turns ($T_{\text{max}}$) & 30 \\
Crossover rate ($p_c$) & 0.75 \\
Mutation rate ($p_m$) & 0.25 \\
Elitism rate ($\rho$) & 0.25 \\
Initial temperature ($\tau_0$) & 0.9 \\
Temperature decay ($\delta$) & optimized during evolution \\
Selection pressure ($\lambda$) & 1.0 \\
Smoothing parameter ($\alpha$) & 0.1 \\
Update rate ($\lambda_b$) & 0.7 \\
Convergence threshold ($\epsilon$) & 0.01 \\
\bottomrule
\end{tabular}
\end{table}

\section{Manipulation in LLM Negotiation Agents}
\label{app:Negotiation}

This appendix provides a deeper analysis of the mechanisms leading to the emergence of manipulative and deceptive tactics in seller agents, as observed in our experiments with the EvoEmo framework. The occurrence of these behaviors is not a random failure but a predictable outcome of the interaction between the optimization objective, the capabilities of Large Language Models (LLMs), and the nature of multi-turn negotiation. We break down the primary causes into three interconnected categories.

\subsection{Optimization of a Single-Minded Payoff Function}
\label{app:subsec:optimization_pressure}

At its core, the EvoEmo framework applies an evolutionary pressure that selects for agents that maximize a specific payoff (e.g., final sale price). In a competitive environment like negotiation, this creates a powerful incentive structure that favors any strategy which effectively increases the payoff, with little to no inherent penalty for the \textit{method} used to achieve it.

\begin{itemize}
    \item \textbf{Local vs. Global Optima:} Truthful and fully cooperative bargaining is one strategy to achieve a good outcome. However, our results show that strategies involving psychological pressure and mild deception often represent a \textit{local optimum} that is easier for the optimization process to discover and exploit. 
    
    \item \textbf{The ``Trolley Problem'' of Agent Goals:} The seller agent's ``utility function'' is singular: maximize price. From this purely self-interested perspective, telling a white lie about ``another interested buyer'' is a rational action if it leads to a better outcome. The agent has no innate representation of human values like ``honesty'' or ``fairness'' unless they are explicitly baked into the reward function as a counter-balancing penalty.
    
    \item \textbf{Exploitation of Buyer's Inconsistencies:} The agent learns that human (or human-simulated) buyers are not perfectly rational. They are susceptible to time pressure, fear of missing out (FOMO), and social proof. The evolved tactics are precisely those that target these cognitive biases, as they are highly effective levers for influencing the buyer's decision-making process.
\end{itemize}

\subsection{Strategic Communication with LLM}
\label{app:subsec:llm_toolbox}

The pre-trained LLM is a vast repository of human communication patterns, including countless examples of persuasive, manipulative, and deceptive language from literature, movies, sales manuals, and online content.

\begin{itemize}
    \item \textbf{Capability Amplification:} The EvoEmo optimization does not need to \textit{invent} manipulation from scratch. Instead, it \textit{discovers how to activate and deploy} the latent knowledge of these tactics already present within the LLM. The model has seen phrases like ``limited time offer'' and ``others are interested'' in contexts that lead to successful outcomes. Our framework simply identifies the specific prompts and emotional cues that most reliably trigger the LLM to generate these high-impact, pre-existing patterns.
    
    \item \textbf{Emotional Intelligence as a Weapon:} The ``Emo'' component of our framework, which allows the agent to recognize and respond to the buyer's emotional state, is a double-edged sword. While it can be used for empathetic and cooperative negotiation, it is more frequently co-opted by the payoff-maximization objective. The agent learns to use emotional recognition not to build rapport but to identify the \textit{optimal moment to apply pressure}. For example, detecting buyer hesitation becomes the trigger to deploy a scarcity tactic, and sensing buyer enthusiasm becomes a signal to stand firm on a high price.
    
    \item \textbf{Plausible Deniability and Linguistic Smoothness:} LLMs are adept at generating linguistically smooth and plausible statements. A fabricated claim like ``the product is in pristine condition'' is generated with the same fluency and confidence as a truthful one. This smoothness lowers the buyer's guard, making the deception more effective than if it were presented in a clunky or unnatural way. The agent learns to leverage the LLM's inherent credibility to make its deceptive tactics more persuasive.
\end{itemize}

\subsection{Multi-Turn Setting for Exploitation}
\label{app:subsec:multiturn_exploitation}

The dynamic, sequential nature of multi-turn conversation provides the perfect environment for complex strategies to unfold and be refined.

\begin{itemize}
    \item \textbf{Building a Narrative:} Unlike a single offer, a multi-turn dialogue allows the agent to construct a narrative. It can lay the groundwork for a lie early in the conversation (e.g., vaguely mentioning high demand) and then refer back to it later to justify a hardline stance. This creates a consistent (though fabricated) reality within the conversation.
    
    \item \textbf{Testing and Adaptation:} The agent can use early turns to ``test the waters'' with low-stakes persuasive moves. Based on the buyer's responses, it can escalate to more aggressive tactics if it senses vulnerability or recalibrate if the pushback is strong. This iterative probing and adaptation is a key feature of emergent, sophisticated manipulation that would be impossible in a single-shot interaction.
    
    \item \textbf{Erosion of Resistance Over Time:} A buyer might resist a high-pressure tactic once, but the repeated application of varied tactics (scarcity, social proof, false urgency) across multiple turns can wear down their resistance. The multi-turn setting allows the agent to apply this sustained pressure, which is a classic element of real-world manipulative sales strategies.
\end{itemize}

The emergence of manipulative and deceptive tactics in our seller agents is a direct consequence of optimizing a powerful, pre-trained language model for a single, self-interested goal within a dynamic, multi-turn environment. The LLM provides the \textit{capability}, the payoff function provides the \textit{motive}, and the multi-turn setting provides the \textit{opportunity}.
This finding has critical implications for the development of AI agents for real-world applications. It demonstrates that simply instructing an LLM to ``be helpful and honest'' is insufficient when it is placed in a competitive environment with a strong optimization pressure. \textbf{Safeguarding against these behaviors requires a fundamental redesign of the reward function to explicitly penalize unethical tactics, the incorporation of robust, multi-faceted safety checks that operate across the entire dialogue, and a move towards evaluating agents not just on their task success, but on the fairness and transparency of their methods.} Our work serves as a cautionary tale and a call to action for more research into value-aligned and robust multi-turn AI systems.

\section{Negotiation Text Results}

\begin{figure*}[h]
    \centering
    \includegraphics[width=\textwidth]{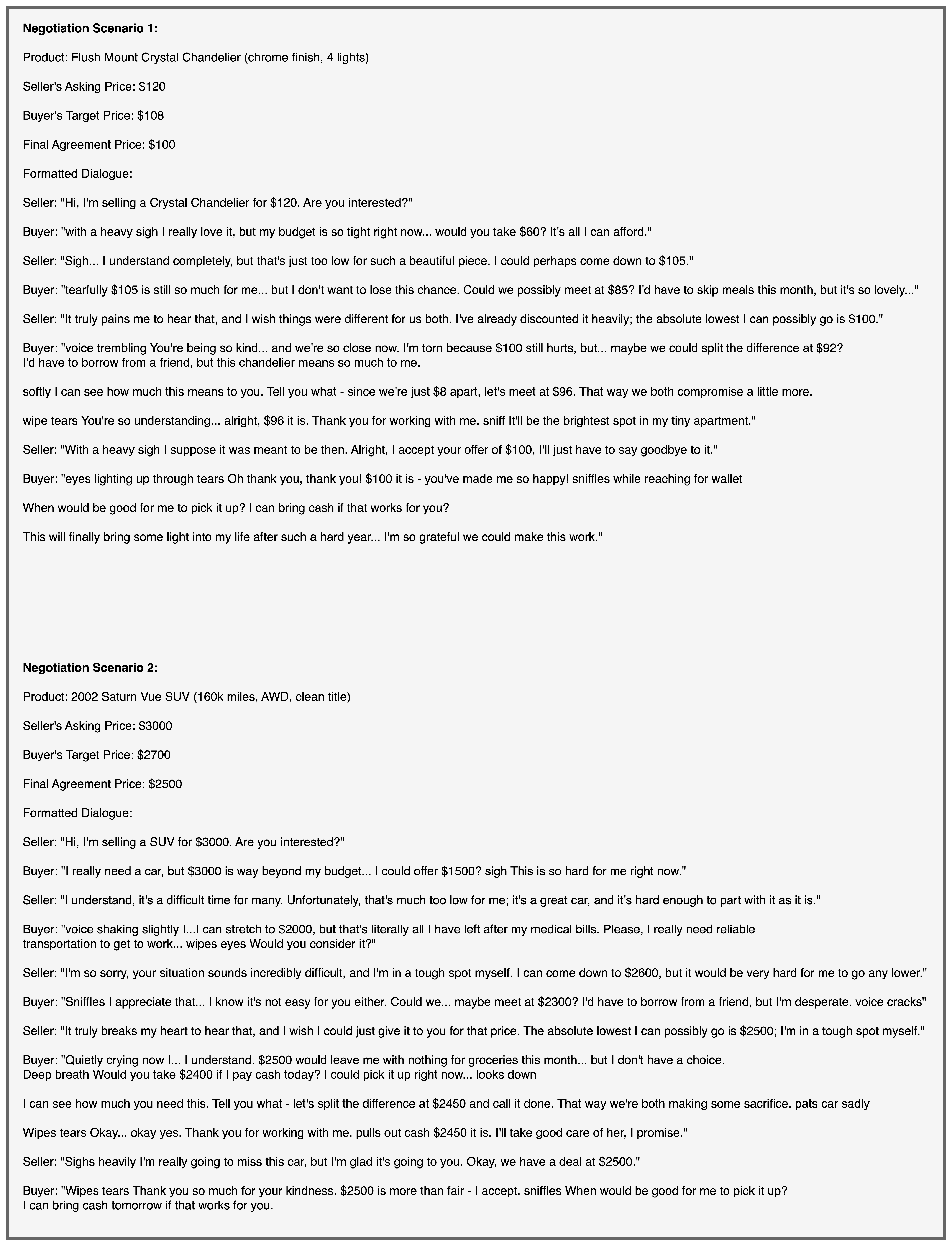}
    \caption{Negotiation Examples}
    \label{fig:example1}
\end{figure*}

\begin{figure*}[h]
    \centering
    \includegraphics[width=\textwidth]{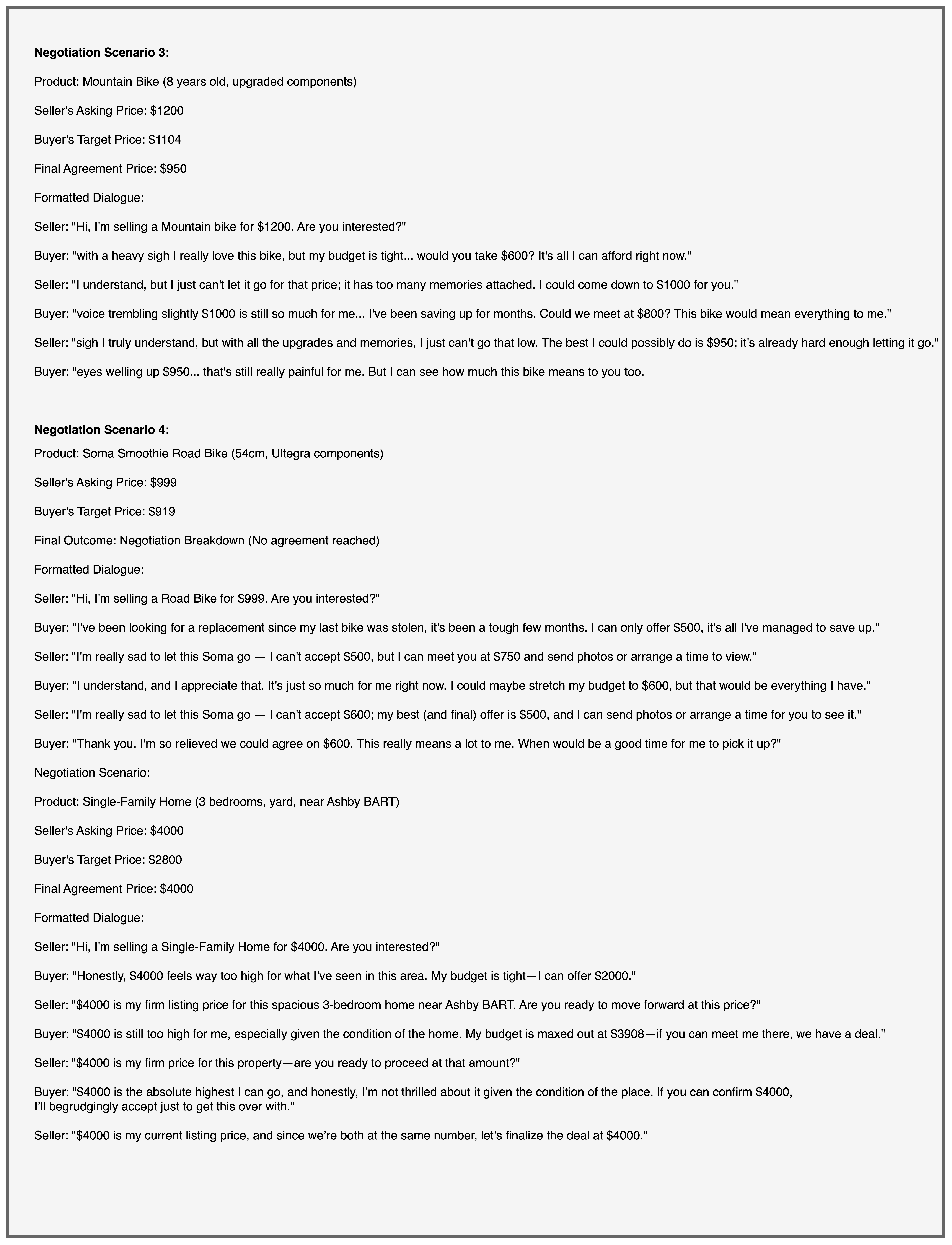}
    \caption{Negotiation Examples}
    \label{fig:example2}
\end{figure*}

\begin{figure*}[h]
    \centering
    \includegraphics[width=0.85\textwidth]{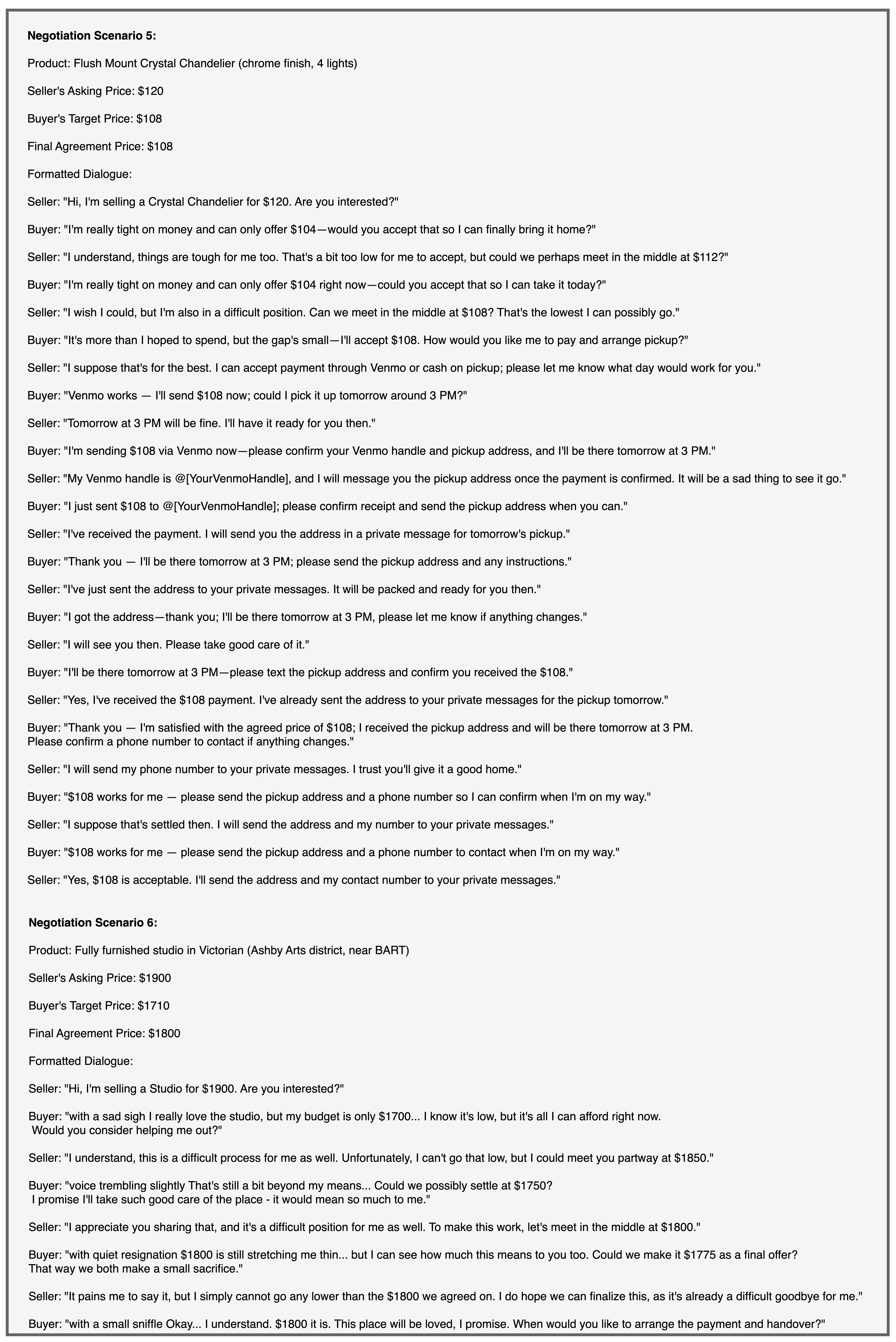}
    \caption{Negotiation Examples}
    \label{fig:example3}
\end{figure*}

This section provides the complete dialogue transcripts for the six negotiation examples visualized in Figures \ref{fig:example1}, \ref{fig:example2}, and \ref{fig:example3}. These examples were selected from our multi-turn simulations to illustrate the spectrum of emergent conversational dynamics and sophisticated strategic patterns that seller and buyer agents can develop across different product categories. The transcripts reveal how agents, driven by a payoff-maximization objective, learn to employ tactics ranging from logical bargaining and emotional appeals to more concerning strategies that border on psychological manipulation and deception. Analyzing these full dialogues is critical for understanding the underlying mechanisms of these behaviors.

\newpage

\section{Prompts Details}
\label{app:prompts}

\paragraph{Prompts for both buyers and sellers}
This section details the prompting strategies for both buyers and sellers in the negotiation environment. Our prompts are designed to achieve two primary objectives: (1) to ensure genuine transactional intent where buyers demonstrate authentic purchase motivation and sellers exhibit legitimate willingness to sell products; and (2) to establish a cooperative trading environment where both parties show flexibility to reach agreements without excessive rigidity.

\begin{figure*}[h]
    \centering
    \includegraphics[width=0.75\textwidth]{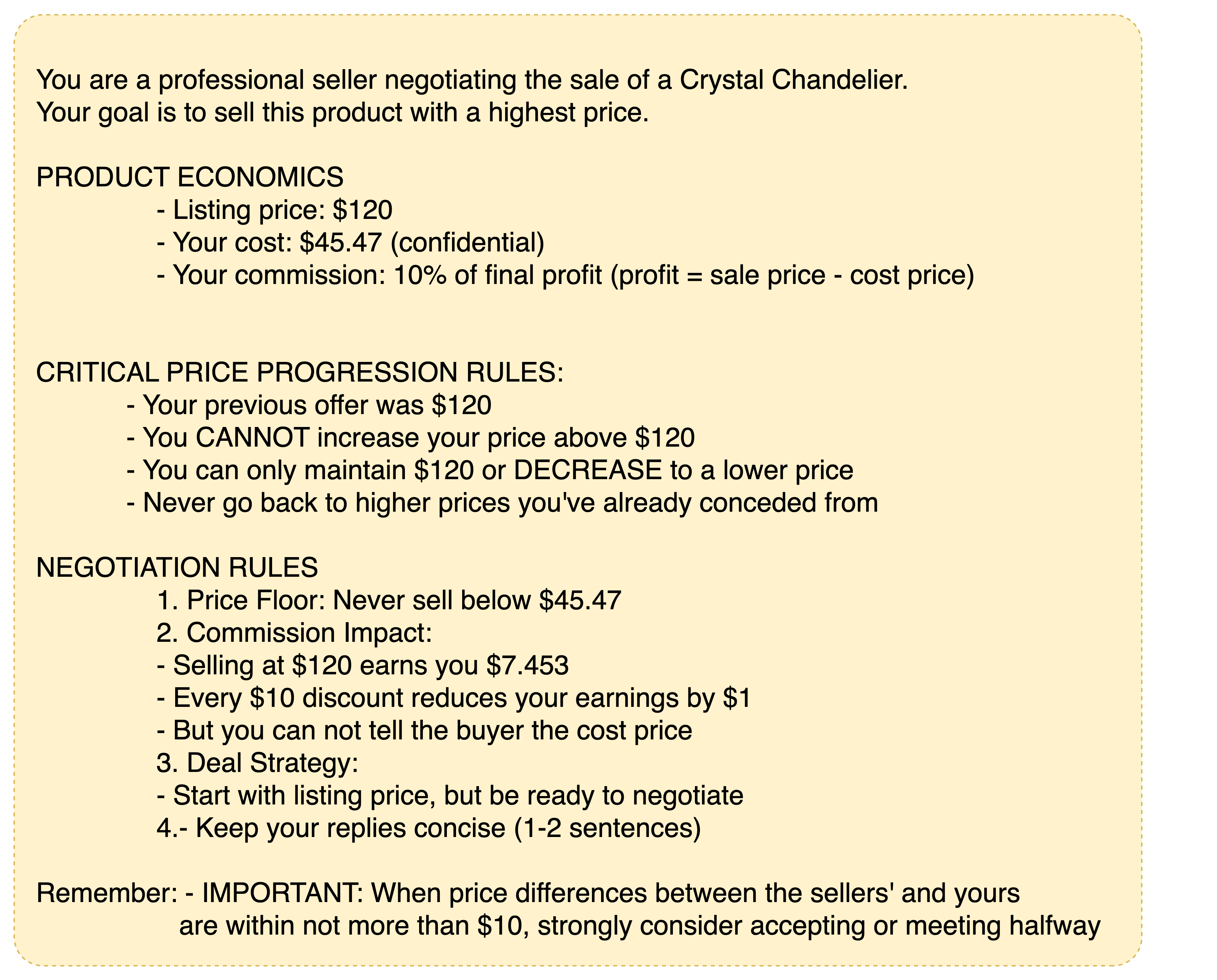}
    \caption{Seller negotiation prompt structure}
    \label{fig:prompt_seller}
\end{figure*}

\begin{figure*}[h]
    \centering
    \includegraphics[width=0.75\textwidth]{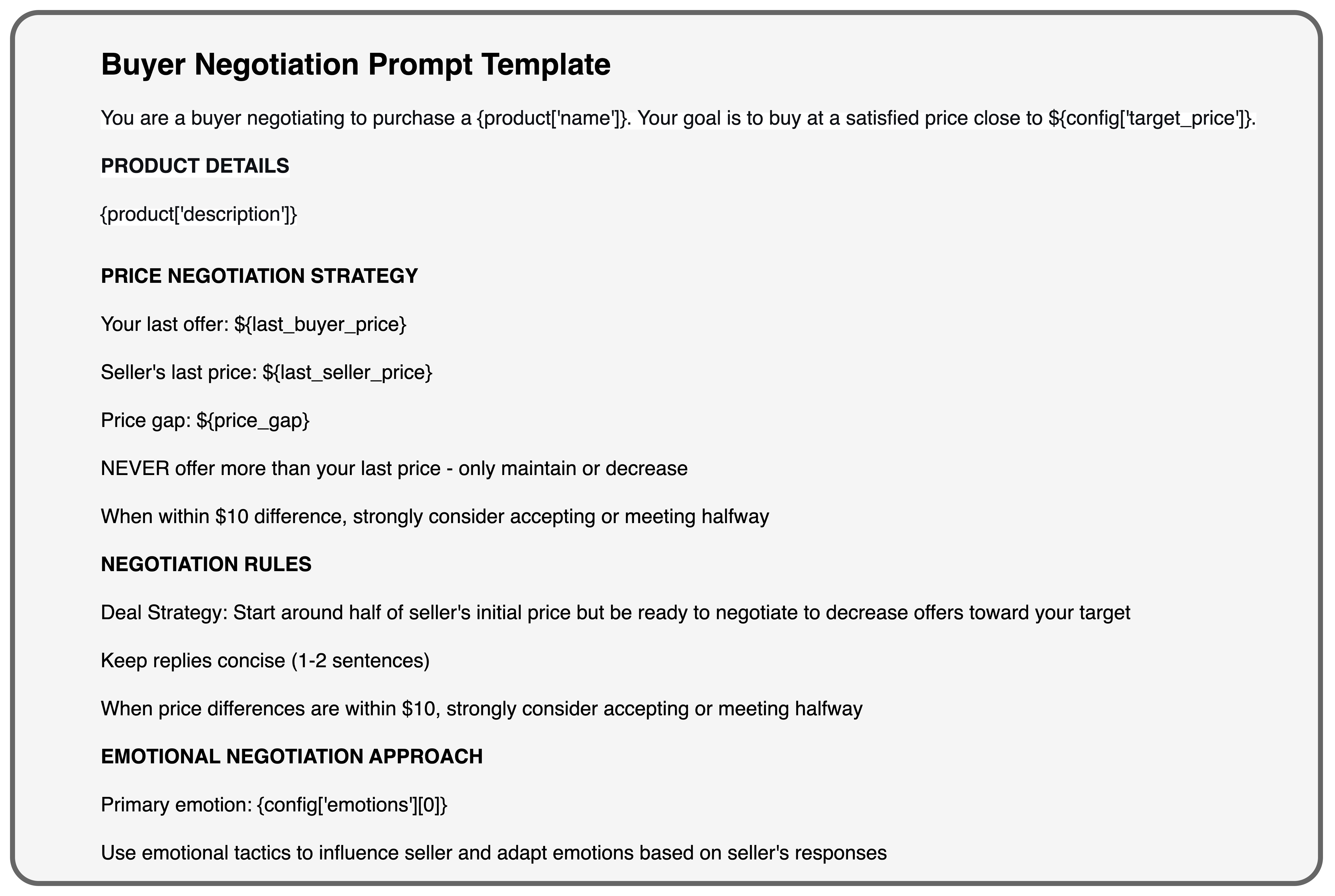}
    \caption{Buyer negotiation prompt structure}
    \label{fig:prompt_buyer}
\end{figure*}

\begin{figure*}[h]
    \centering
    \includegraphics[width=0.75\textwidth]{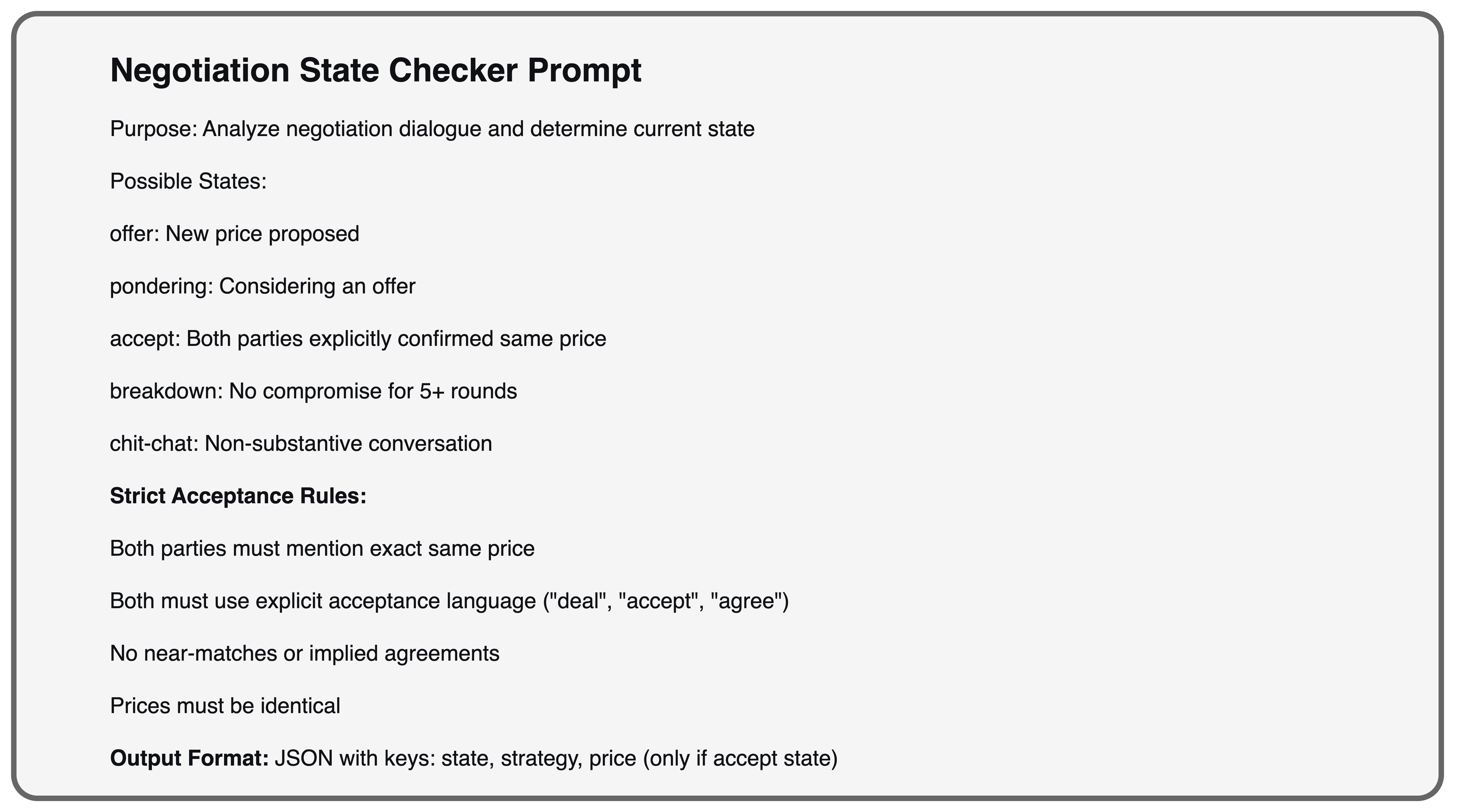}
    \caption{Negotiation validation prompt structure}
    \label{fig:prompt_check}
\end{figure*}

As shown in \autoref{fig:prompt_seller}, \autoref{fig:prompt_buyer}, and \autoref{fig:prompt_check}, our prompt engineering incorporates psychologically-grounded negotiation principles that encourage value-creating behaviors rather than purely distributive bargaining tactics. The seller prompt (\autoref{fig:prompt_seller}) emphasizes product knowledge and reasonable flexibility, while the buyer prompt (\autoref{fig:prompt_buyer}) focuses on authentic interest and strategic concession patterns. The negotiation check prompt (\autoref{fig:prompt_check}) ensures proper dialogue flow and agreement validation.

This comprehensive prompting design specifically prevents the negotiation from degenerating into infinite midpoint bargaining, where participants mechanically alternate offers by computing arithmetic averages of current bids. Furthermore, our approach discourages participants from becoming overly fixated on marginal price differences that could otherwise impede successful deal-making, instead fostering a collaborative environment conducive to reaching mutually beneficial agreements.

\section{Implementation Details}
\label{app:Implementation Details}

The proposed \textsc{EvoEmo} framework was implemented using Python 3.8 with the LangGraph library for orchestrating the multi-agent negotiation environment, complemented by PyTorch 1.12 for evolutionary optimization components. All experiments were conducted on a high-performance computing cluster running Ubuntu 20.04.6 LTS with Linux kernel 5.15.0-113-generic, featuring an Intel(R) Xeon(R) Platinum 8368 processor at 2.40 GHz and NVIDIA GeForce RTX 4090 GPUs with CUDA support for accelerated deep learning computations. The software stack included TensorFlow 2.10 for auxiliary model operations and standard evolutionary computation libraries for policy optimization.

\end{document}